\documentclass[letterpaper, 10 pt, conference]{IEEEconf}  

\IEEEoverridecommandlockouts                              
\usepackage{graphicx} 
\usepackage[absolute,overlay]{textpos}
\usepackage{xcolor} 
\usepackage{float}
\usepackage{pifont} 
\usepackage{amssymb}
\usepackage{amsmath}
\usepackage{multirow}
\usepackage{algorithm}
\usepackage{algpseudocode}

\usepackage{amsthm}

\let\labelindent\relax
\usepackage{enumitem}
\usepackage{array}
\usepackage{booktabs}
\usepackage{caption}
\usepackage{subcaption}

\title{\LARGE \bf
Communication-Constrained Multi-Robot Exploration With Adaptive Communication Windows
}

\author{Ben Rossano$^{1,3}$, Jaein Lim$^2$, Jonathan P. How$^1$
\thanks{$^{1}$B. Rossano and J.P. How are with the Aerospace Controls Lab, Massachusetts Institute of Technology, Cambridge, MA, USA {\tt\small \{brossano, jhow\}@mit.edu}}%
\thanks{$^{2}$J. Lim is with the Charles Stark Draper Laboratory, Cambridge, MA, USA {\tt\small {jlim}@draper.com}}%
\thanks{$^3$ B. Rossano is a Draper Scholar with the Charles Stark Draper Laboratory, Cambridge, MA. The authors would like to thank the Draper Scholars program for supporting this work.}
}

\usepackage[noadjust,sort,compress]{cite}

\makeatletter
\let\NAT@parse\undefined
\makeatother

\usepackage[colorlinks=true,linkcolor=black,citecolor=black,urlcolor=blue]{hyperref}

\makeatletter
\let\orglabel\label
\renewcommand{\label}[1]{\orglabel{#1}\hypertarget{#1}{}}
\makeatother

\begin{document}

\setlength{\textfloatsep}{5pt}

\maketitle
\thispagestyle{empty}
\pagestyle{empty}
\begin{abstract}
Exploring unknown environments with multi-robot teams can improve efficiency by allowing robots to explore in parallel. However, realizing these gains requires effective information sharing. When communication is intermittent, robots must balance the benefits of sharing information against the cost of diverting from exploration to establish communication. This paper introduces MACE, a decentralized exploration framework that actively evaluates whether establishing communication is worthwhile. At scheduled communication windows, robots estimate the cost of reaching previously identified communication locations. By formulating this decision as a variant of the Vehicle Orienteering Problem, robots evaluate routes based on the travel required to establish communication and the exploration that can be completed along the way. This approach enables robots to communicate more frequently than under purely opportunistic strategies while reducing the unnecessary travel associated with fixed rendezvous strategies. Across a set of simulated environments with varying size and geometry, we demonstrate that MACE reduces the total exploration time by up to 23\% compared to existing communication-constrained exploration strategies.
\end{abstract}

\section{Introduction} \label{sec:introduction}

The goal of autonomous exploration is to navigate through an unknown environment to maximally gain information, either within a fixed time or until complete coverage \cite{Ericson_time_vs_gain}. Multi-robot teams are particularly well-suited for these tasks as they can parallelize the exploration process, achieving efficiency gains that single robots cannot match. Realizing these gains, however, requires intelligent coordination strategies to avoid redundant coverage, which occurs when a robot unknowingly re-explores an area that a teammate has already covered. 

One prominent line of work assumes that robots maintain continuous communication, allowing each robot to access the latest information available to the entire team and enabling exploration in a fully coordinated fashion \cite{Dutta_fully_connected}. While these approaches yield high exploration efficiency, the assumption of constant, reliable communication frequently breaks down in practice. In environments without global network infrastructure, such as tunnel systems, urban areas experiencing communication outages, and unstructured outdoor terrain, robots must instead form mobile ad-hoc networks (MANETs) to communicate directly with one another \cite{clark2022propem}. However, limited mobile radio range and signal attenuation through materials such as concrete and metal make continuous team-wide connectivity difficult to sustain without significantly constraining coverage.

\begin{figure}[t]
    \centering
    \includegraphics[width=0.48\textwidth]{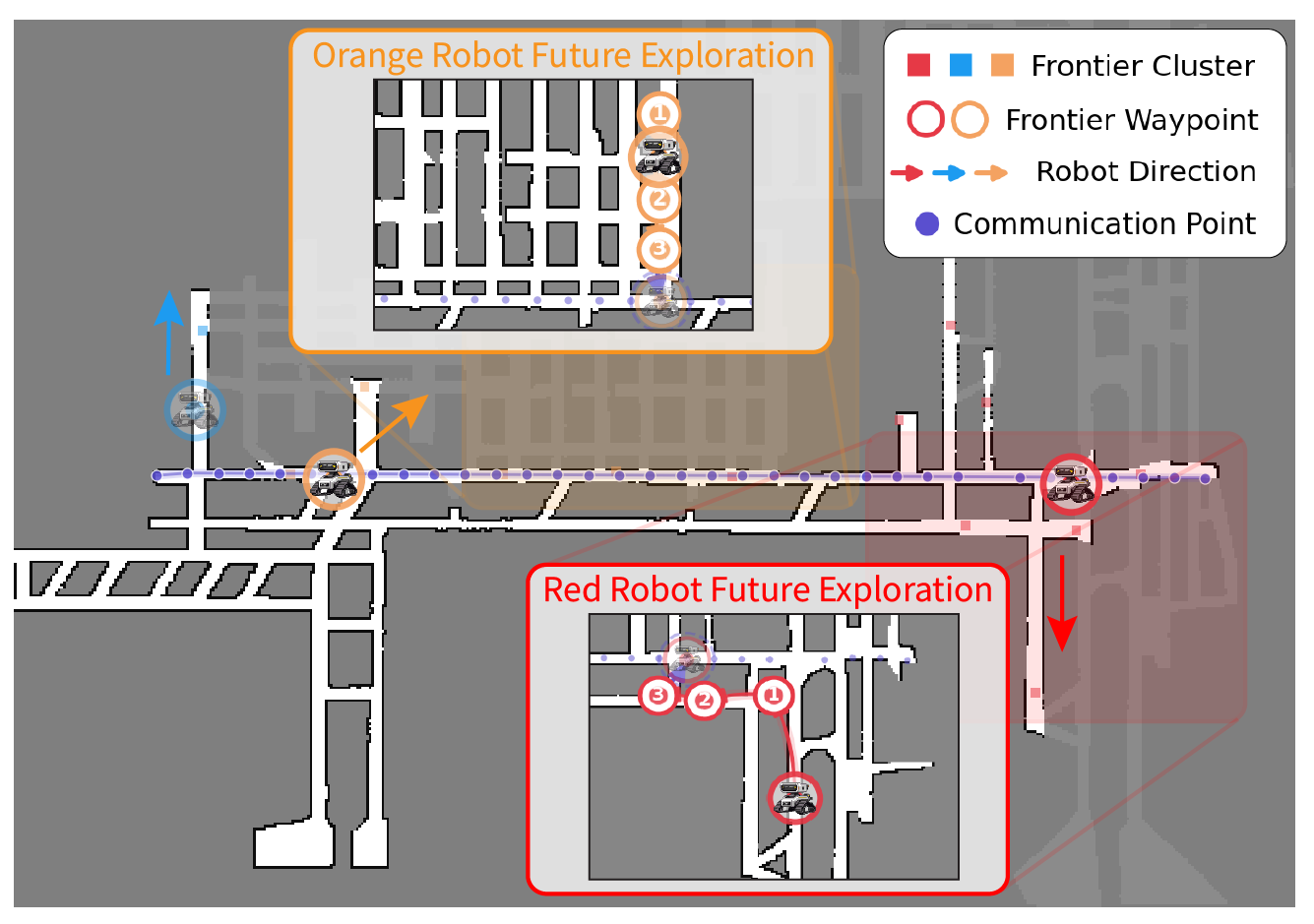}
    \caption[Overview of MACE's communication-aware exploration strategy]{Overview of MACE's communication-aware exploration strategy. When robots establish communication, they update points for future communication. Later on, the robots determine the cost of traveling back to these points through a route of frontiers to establish communication.}
    \label{fig:cover_exploration}
\end{figure}

As a result, multi-robot exploration under intermittent communication has emerged as an important research challenge, motivating strategies that enable effective exploration without constant connectivity \cite{comm_restricted_survey}. Existing work falls broadly into two categories. \textit{Rendezvous-based strategies} designate specific times and locations for robots to meet and share information, actively enforcing map consistency to mitigate redundant exploration \cite{dasilva_rendezvous}. However, rendezvous locations must be selected using limited knowledge of the environment and may become poorly positioned as exploration progresses, forcing robots to make long detours through previously-explored space.

On the other hand, \textit{opportunistic strategies} do not coordinate meetings. Instead, robots share information only when they happen to encounter one another, relying on information from prior contacts to guide their exploration policies in between encounters \cite{racer}. While this avoids the shortcomings of mandatory meetings, robots that go long periods without contact are more prone to redundant exploration. The limitations of these strategies motivate a solution approach that more intelligently trades off exploration progress and information exchange.  

In this work, we present MACE (\underline{M}ulti-robot  \underline{A}daptive \underline{C}ommunication-constrained \underline{E}xploration), a decentralized framework that combines scheduled communication windows with intelligent evaluation of whether communication is worth pursuing. During a communication window, rather than committing to a rendezvous, each robot performs a rollout to estimate the cost of establishing communication with its teammates based on previously determined communication locations. By formulating this as a variant of the Vehicle Orienteering Problem (VOP), robots estimate the ``detour cost" of attempting communication---specifically, how far away a potential meeting point is and how much exploration can be conducted en route to it. If this cost is low, the robot attempts to establish communication. Otherwise, it continues exploring and re-evaluates at the next window. This approach preserves the structure of rendezvous-based strategies by scheduling dedicated opportunities for communication, but removes the requirement that robots attend meetings regardless of cost. Since communication attempts are only made when the cost is low, their impact on exploration efficiency is limited, while the regularly scheduled communication windows produce more frequent information exchange than purely opportunistic methods. Our work can be summarized by the following contributions: 

\begin{itemize}
\item We propose MACE, a decentralized multi-robot exploration framework that balances exploration efficiency with active communication awareness under intermittent connectivity constraints.
\item We evaluate MACE against two baseline approaches across a set of simulated environments, ranging in size and geometry. Our results demonstrate that MACE's communication-aware exploration approach can decrease the total time needed to explore an environment by up to 23\%.
\item We analyze how environment size and geometry influence the performance of different exploration strategies using multiple graph-based connectivity metrics.
\end{itemize}

\section{Background and Related Work} \label{sec:rel_work_mre}

\subsection{Frontier Exploration}
Frontier exploration is the process of traveling to boundaries between known and unknown space to increase a robot's knowledge of the environment \cite{yamauchi}. The primary challenge of this process is determining which frontiers to visit to maximize exploration efficiency. Previous work has extensively investigated different methods for defining the utility of a frontier, including proximity to the robot, expected information gain, and spread relative to teammates \cite{review_utility}.

One common frontier selection approach is to greedily assign each robot the highest-scoring available frontier, replanning each time a new frontier is reached with updated information \cite{faigl2013greedy_comp}. While this enables quick replanning, it can be susceptible to the standard myopic behavior of greedy planning. Other approaches instead plan sequences of frontiers rather than individual targets \cite{cao2021tare}, \cite{cao2023mtare}. However, planning sequences of frontiers does not always provide additional value, since the frontier set changes continuously as new regions are discovered through exploration.

\subsection{Connectivity-Constrained Methods}
Connectivity-constrained approaches maintain team-wide connectivity throughout the mission, ensuring that information is shared continuously \cite{Dutta_fully_connected, Pratissoli_connected_coverage, bautin2012minpos}. While this eliminates redundant exploration, it can reduce efficiency by limiting how spread out the team explores, especially in large or geometrically complex environments. Despite these limitations, connectivity-constrained approaches introduce solution techniques that remain relevant in our setting. In particular, the problem of positioning a multi-robot team to satisfy connectivity can be reframed to identify good meeting locations for a disconnected team. In \cite{scherer2020multi, kantaros2016global}, this problem is formalized using connected subgraph optimization, enabling existing graph solvers to identify locations that satisfy communication constraints. 

\subsection{Rendezvous-Based Methods}
Rendezvous-based approaches schedule designated times for robots to meet and exchange information, guaranteeing that robots share information at regular intervals. Some methods require the full team to rendezvous at each scheduled event \cite{hollinger_periodic}, while others optimize a schedule of pairwise or subteam meetings, allowing some robots to continue exploring while others communicate \cite{dasilva_rendezvous, meet_merge}.

Scheduled communication provides a reliable mechanism for information exchange but introduces a fundamental planning challenge: at the time of each meeting, robots must select a location for the next rendezvous without knowing where exploration will take them in the interim. Common rendezvous selection approaches, such as the center of mass of current frontiers \cite{bramblett_rendezvous} or a randomly sampled frontier \cite{dasilva_rendezvous}, tend to become poor strategies by the time the next rendezvous arrives, as robots push deeper into unknown space and the frontier boundaries shift substantially. Recent work leverages learned map predictions to make more informed estimates of what lies beyond current frontiers, which could prove to be useful for rendezvous selection \cite{mapex}.

\subsection{Opportunistic Methods}
Opportunistic approaches do not rely on fixed meeting schedules, instead allowing robots to share information whenever they encounter one another by chance \cite{racer, mocha}. This avoids the backtracking costs inherent to rendezvous-based methods, since robots are not regularly diverting from exploration to satisfy pre-committed meetings. However, without any mechanism to enforce encounters, robots may go extended periods without communicating, which can lead to redundant exploration.

Recent work has introduced pursuit-based communication \cite{cao2023mtare}, in which a robot can actively seek out teammates using its last known estimate of their positions and intended plans. While this may increase the likelihood of communication, stale knowledge can also be a poor predictor of a robot's actual location, particularly as robots push further into unknown space. Thus, pursuit may also \textit{promote} redundant exploration in certain envirornments.

\section{Problem Formulation}
\label{sec:problem}

We consider a team of $N$ robots $\mathcal{R} = \{1, 2, \dots, N\}$ tasked with
collaboratively exploring an unknown bounded environment $\mathcal{W} \subset
\mathbb{R}^2$. Each robot $i \in \mathcal{R}$ maintains an occupancy-grid map
$m_i : \mathcal{W} \to \{\text{free},\, \text{occupied},\, \text{unknown}\}$
that it updates from onboard range measurements as it moves. The frontier set
$\mathcal{F}_i \subset \mathcal{W}$ is defined as the boundary between free
and unknown cells in $m_i$.  For planning purposes, each frontier $f \in \mathcal{F}_i$ is represented by a single 2D coordinate, chosen as the location of the free-space cell closest to the frontier's centroid. The team objective is to cover the explorable portion of $\mathcal{W}$ in the minimum cumulative time.

When two robots are in communication, they exchange their local maps and share their future intentions. We assume that each robot can localize itself perfectly within the environment and accurately map its surroundings. Under these assumptions, map fusion becomes straightforward as the local maps can be directly aligned and merged, with priority given to occupied cells. Furthermore, each robot $i$ maintains, for every other robot $j$, a plan $p^{(i)}_j = (\hat{f}^{(i)}_j,\, \hat{q}^{(i)}_j,\, t^{(i)}_j)$ storing the next frontier $\hat{f}$ that robot $j$ intends to pursue, a larger region $\hat{q}$ that robot $j$ has been assigned to explore, and the timestamp $t$ at which this information was received. This allows robots to propagate knowledge about one another through indirect encounters, always preferring the most recent information.

Each robot also stores a set of communication points $\mathcal{C}_{ij}$, which are locations such that communication between robots $i$ and $j$ is guaranteed if both robots are present simultaneously. Formally, for a pair $(i, j)$ and some communication model, we seek a set of points $\mathcal{C}_{ij} \subset \mathcal{W}$ such that
\begin{equation}
\label{eq:corridor_def}
\forall\, w_a, w_b \in \mathcal{C}_{ij}, \quad \texttt{inComms}(w_a, w_b) = \text{true},
\end{equation}
where $\texttt{inComms}()$ returns true if a pair of points satisfies the communication model (e.g., line-of-sight, disk). Thus, if robot $i$ travels to any point in $\mathcal{C}_{ij}$ at some scheduled communication time $t^{c}_{ij}$, and robot $j$ does the same, communication will be re-established. We introduce two additional parameters that influence opportunities for communication: (i) the communication interval $\Delta_c$ governs how long after the current encounter the next $t^{c}_{ij}$ is scheduled, and (ii) the lookahead horizon $\delta$ controls how soon before $t^{c}_{ij}$ robot $i$ begins its communication-aware planning (Section \ref{subsec:comm_aware}).

\section{Method}
MACE consists of three primary components: default exploration, communication point updates, and communication-aware exploration. Algorithm \ref{alg:whole_explore_strat} outlines the high-level planning strategy executed by a single robot $i$. At each timestep, if communication with any teammate is available, the robots exchange local information, including maps and planned intentions. If the change in the map since the previous update exceeds a threshold $\Delta m$, the robots update their assigned high-level exploration regions (Algorithm \ref{alg:connected_assignment}), advance the next communication opportunity time $t^{c}_{ij}$, and update $\mathcal{C}_{ij}$ (Algorithm \ref{alg:update_comm_points}).

\begin{algorithm}[t]
\caption{Robot $i$ Exploration Strategy}
\label{alg:whole_explore_strat}
\begin{algorithmic}[1]
\small
\State $t \gets 0$
\State $m_i, m_{\text{prev}}, \tau_i, \pi_i \gets \emptyset, \emptyset, \emptyset, \emptyset$
\State $t^{c}_{ij} \gets \Delta_c,\ \text{missed}_{ij}\gets0, \ \mathcal{C}_{ij} \gets \emptyset,\ p^{(i)}_{j} \gets \emptyset \quad \forall j \in \mathcal{R}\setminus\{i\}$
\State $\mathcal{F}_i \gets$ \Call{findFrontiers}{$m_i$}
\While{$|\mathcal{F}_i| > 0$}
    \State $\mathcal{S} \gets  \bigl\{\, j \;:\; j \in \mathcal{R} \setminus \{i\}, \ $\Call{inComms}{$\text{pos}_i, \text{pos}_j$}$\,\bigr\}$
    \If{$|\mathcal{S}| > 0$}
        \State $m_i,\ p^{(i)} \gets$ \Call{fuseInfo}{$\mathcal{S}$}
        \If {\Call{Diff}{$m_i, m_{\text{prev}}$} $> \Delta m$}
            \State $\mathcal{F}_i \gets$ \Call{findFrontiers}{$m_i$}
            \State $p^{(i)}\gets $ \Call{UpdateRegions}{$\mathcal{F}_i$, $p^{(i)}$, $\mathcal{S}$, $i$}
            \State $t^{c}_{ij} \gets t + \Delta_c,\ \text{missed}_{ij}\gets 0 \quad \forall j \in \mathcal{S}$ 
            \State $\mathcal{C}_{ij} \gets$ \Call{updateCommPoints}{$\mathcal{F}_i, p^{(i)}, i, j$} $\quad \forall j \in \mathcal{S}$
            \State $m_{\text{prev}} \gets m_i, \ \tau_i \gets \emptyset$
        \EndIf
    \EndIf
     
    \If{$\tau_i = \emptyset$}
        \State $\mathcal{F}_i \gets$ \Call{findFrontiers}{$m_i$}
        \If{$\exists\, j : t > t^{c}_{ij} - \delta$}
            \State $\sigma \gets$ \Call{solveOrienteering}{$\mathcal{F}_i, \mathcal{C}_{ij}, t^{c}_{ij} - t$}
            \If{$\sigma$ valid}
                \State $\tau_i \gets \sigma_0$
            \EndIf
        \EndIf
        \If{$\tau_i = \emptyset$}
            \If {\Call{closestFrontier}{$\mathcal{F}_i, p^{(i)}$} $ > d_f$}
                \State $p^{(i)}\gets $ \Call{UpdateRegions}{$\mathcal{F}_i$, $p^{(i)}$, $\{i\}$, $i$}
            \EndIf
            \State $\tau_i \gets \arg\max_{f \in \mathcal{F}_i}$ $U_{\text{front}_i}(f, \hat{f}^{(i)})$
        \EndIf
        \State $\pi_i \gets$ \Call{AStar}{$\text{pos}_i, \tau_i, m_i$}
    \EndIf
    \If{$t > t^{c}_{ij}$}
        \State $t^{c}_{ij} \gets t + \Delta_c$
        \State $\text{missed}_{ij}\gets \text{missed}_{ij}+1$
    \EndIf
    \State $\text{pos}_i \gets$ \Call{FollowPath}{$\pi_i$}
    \If{$\text{pos}_i = \tau_i$}
        \State $\tau_i \gets \emptyset$
    \EndIf
    \State $t \gets t+1$
\EndWhile
\end{algorithmic}
\end{algorithm}

Whenever a new frontier target is required, either because the robot has reached its current target $\tau_i$ or because the exploration regions have been updated, the robot checks whether any teammate’s scheduled communication opportunity occurs within the time horizon $\delta$. If such an opportunity exists, the robot selects the teammate that it has been out of communication with the longest and invokes the communication-aware orienteering solver (Section \ref{subsec:comm_aware}). If a feasible route exists, this solver returns a chain $\sigma$ of frontier waypoints terminating at a point in $\mathcal{C}_{ij}$, subject to the remaining time budget $t^{c}_{ij}-t$, and the first waypoint $\sigma_0$ becomes the new frontier target. If the solver does not return a feasible chain, the robot checks whether there are any feasible communication opportunities with other robots. When no communication-aware window is active or the solver does not find any feasible chain, the robot falls back to the default exploration strategy. A path $\pi_i$ is then planned with A* search. If the current time surpasses a scheduled communication opportunity, the event is treated as a missed communication opportunity, and the next opportunity is rescheduled to occur $\Delta_c$ time steps later. 

Although not explicitly described in Algorithm \ref{alg:whole_explore_strat}, we implement a mandatory rendezvous fallback mechanism in which robots are forced to rendezvous at $\mathcal{C}_{ij}$ after a large number of missed opportunities. This threshold is offset across different robot pairs to ensure that a robot is not simultaneously required to attend mandatory rendezvous with multiple teammates.

\subsection{Default Exploration}
\label{sec:default_exploration}

Our default exploration consists of a hierarchical strategy where each robot first gets assigned an exploration region and then greedily selects frontier targets given that region.

\subsubsection{Exploration Region Selection}
\label{sec:region_strategy}

Let $q_i$ denote robot $i$'s current exploration region, and let $\mathcal{Q}_i$ be the candidate region set. This set is computed by clustering nearby individual frontiers and finding the centroid of each. Robots may update their exploration regions under two conditions: (i) robots are in communication with other robots and their fused map is sufficiently different from the map used during the previous region update, or (ii) a robot has explored all frontiers within $d_f$ of the region centroid. For a robot $i$ and candidate region $q \in \mathcal{Q}_i$, the utility is defined as:
\begin{equation}
\label{eq:region_score}
\scalebox{0.9}{$U_{\text{region}_i}(q, \hat{q}^{(i)}) \;=\; -\,\alpha_d \, d_i(q) \;+\; \alpha_g \, g_i(q) \;+\; \alpha_s \, s_i(q, \hat{q}^{(i)}),$}
\end{equation}

where $d_i$, $g_i$, and $s_i$ denote the normalized travel distance, expected information gain, and coverage spread, respectively, and $\alpha_d, \alpha_g, \alpha_s \ge 0$ are their associated weighting coefficients. The travel term is the normalized minimum path distance from the robot's position to the region centroid on the current free-space map, computed for all centroids by a single Dijkstra pass on $m_i$. The information-gain term estimates the unknown area that will be revealed at the frontiers of this region, which is approximated using ray-casting to simulate LiDAR coverage at each frontier location. Lastly, the spread term penalizes regions that lie too close to any region already claimed by another robot, encouraging the team to disperse across the environment:
\begin{equation}
\label{eq:spread_term}
s_i(q, \hat{q}^{(i)}) \;\approx\; \min_{\hat{q}_j \in \hat{q}^{(i)}} \| q - \hat{q}_j \|_2 .
\end{equation}
All distance terms are normalized by the maximum distance over the corresponding candidate set.

When robots are in communication during an exploration region update, a greedy auction over regions is conducted as described in Algorithm \ref{alg:connected_assignment}. This is implemented by electing one robot to compute the solution and broadcast the resulting assignments back to the other robots in $\mathcal{S}$---the subteam of robots currently in communication. 
The previously selected regions for robots in $\mathcal{S}$ are first removed from the set of active regions $\hat{q}^{(\mathcal{S})}$, allowing the communicating robots to revise their assignments using the newly fused information. During each round, the winning robot and region are selected, where $b_k$ denotes robot $k$'s largest bid on any available region. Updating $\hat{q}^{(\mathcal{S})}$ between rounds ensures that subsequent bidders are repelled from already-claimed regions, yielding a spatially spread allocation. Each robot then updates its known plans based on the results of the auction and the current timestamp. When a robot performs an individual update (Algorithm \ref{alg:whole_explore_strat}, Line 27), the procedure is the same, although $\mathcal{S} = \{i\}$, which removes the auction aspect of the selection. 

\begin{algorithm}[t]
\caption{Exploration Region Greedy Assignment}
\label{alg:connected_assignment}
\begin{algorithmic}[1]
\small
\Procedure{UpdateRegions}{$\mathcal{F}_{\mathcal{S}}, p^{(\mathcal{S})},  \mathcal{S}, i$}
    \State $\mathcal{Q}_{\mathcal{S}} \gets$ \Call{mergeFrontierRegions}{$\mathcal{F}_{\mathcal{S}}$}
    \State $\hat{q}^{(\mathcal{S})} \;=\; \bigl\{\, \hat{q}_j^{\,(\mathcal{S})} \;:\; j \in \mathcal{R}, \ \hat{q}_j^{\,(\mathcal{S})} \neq \emptyset \,\bigr\}$
    \State $\hat{q}^{(\mathcal{S})}_k \gets \emptyset \quad \forall k \in \mathcal{S}  $
    \State $\mathcal{S}_{\text{rem}} \gets \mathcal{S}$ 
    \While{$\mathcal{S}_{\text{rem}} \neq \emptyset$}
        \For{$k \in \mathcal{S}_{\text{rem}}$}
            \State $b_k \gets \max_{q \in \mathcal{Q}_\mathcal{S}} U_{\text{region}_k}\bigl(q, \hat{q}^{(\mathcal{S})}\bigr)$
        \EndFor
        \State $k^* \gets \arg\max_{k \in \mathcal{S}_{\text{rem}}} b_k$ 
        \State $q^* \gets \arg\max_{q \in \mathcal{Q}_\mathcal{S}} U_{\text{region}_{k^*}}\bigl(q, \hat{q}^{(\mathcal{S})}\bigr)$
        \State $\hat{q}^{(\mathcal{S})}_{k^*} \gets q^*$
        \State $\mathcal{S}_{\text{rem}} \gets \mathcal{S}_{\text{rem}} \setminus \{k^*\}$
    \EndWhile
    \State $p^{(\mathcal{S})} \gets$ \Call{UpdatePlans}{$\hat{q}^{(\mathcal{S})}$}
    \State \Return $p^{(\mathcal{S})}$
\EndProcedure
\end{algorithmic}
\end{algorithm}

\subsubsection{Frontier Selection}
\label{sec:frontier_given_region}

Once robot $i$ has been assigned an exploration region $q_i$, it must select a specific frontier to visit next. For a  candidate frontier $f \in \mathcal{F}_i$, the utility is defined as:
\begin{equation}
\label{eq:frontier_score}
\begin{aligned}
U_{\text{front}_i}(f, \hat{f}^{(i)}) \;=\;&
-\beta_d \, d_i(f) + \beta_g \, g_i(f) - \beta_r \, r_i(f) \\
&+ \beta_s \, s_i(f, \hat{f}^{(i)}),
\end{aligned}
\end{equation}
where $d_i$, $g_i$, $r_i$, and $s_i$ denote the normalized travel distance, expected information gain, region alignment, and coverage spread, respectively, and $\beta_d, \beta_g, \beta_r, \beta_s \ge 0$ are their associated weighting coefficients. The travel term and information-gain term are defined as in Eq. \eqref{eq:region_score}. The spread term $s_i(f, \hat{f}^{(i)})$ penalizes frontiers near any already-claimed frontiers, with the same form as Eq. \eqref{eq:spread_term}.

The region-alignment term $r_i(f)$ couples frontier selection with the coarse exploration region assignment by comparing two path distances: the robot-to-region distance $d_i(q_i)$ and the frontier-to-region distance $d(f, q_i)$. The quantity $d_i(q_i) - d(f,q_i)$ captures how much visiting $f$ reduces the remaining distance to $q_i$. 
This encourages exploration toward the assigned region without forcing the robot to ignore frontiers encountered along the way. 


\begin{algorithm}[t]
\caption{Updating Communication Points}
\label{alg:update_comm_points}
\begin{algorithmic}[1]
\small
\Procedure{updateCommPoints}{$\mathcal{F}_i, p^{(i)}, i, j$}
    \State $\mathcal{Q} \gets$ \Call{mergeFrontierRegions}{$\mathcal{F}_i$}
    \State $\bar{c}_i, \bar{c}_j \gets$ \Call{clusterAcrossTeam}{$\mathcal{Q}$, $p^{(i)}$}
    \State $G_{\text{move}}, G_{\text{comm}} \gets$ \textsc{GetPlanningGraphs}{()}
    \State $\mathcal{C}_{ij} \gets$ \Call{solveCommPoints}{$G_{\text{comm}}, G_{\text{move}}, \{\bar{c}_i, \bar{c}_j\}$} 
    \State \Return $\mathcal{C}_{ij}$
\EndProcedure
\end{algorithmic}
\end{algorithm}
\begin{figure*}[t]
\centering

\begin{subfigure}[t]{1.4in}
    \centering
    \includegraphics[height=1.45in]{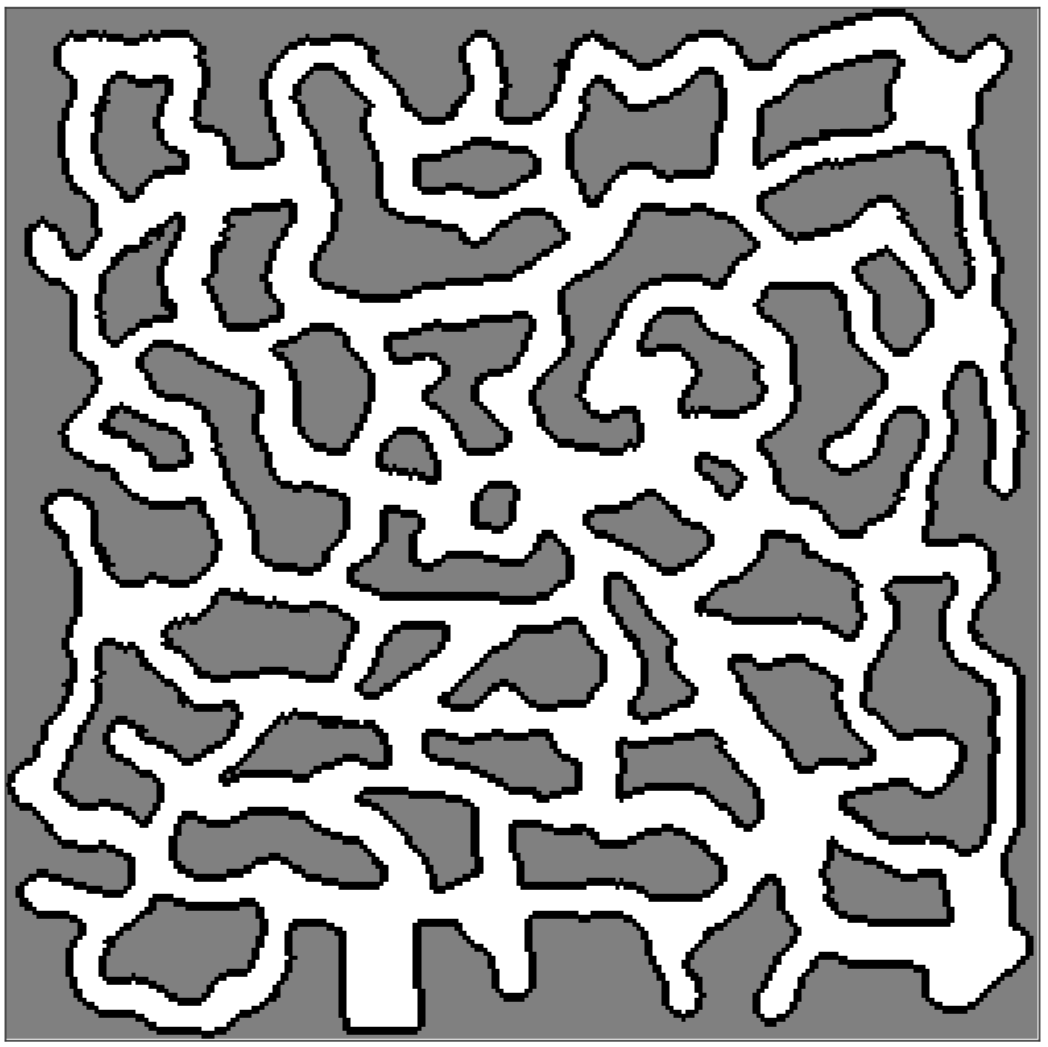}
    \caption{Maze ($250 \times 250$ m)}
    \label{fig:gt_maze}
\end{subfigure}
\hfill
\begin{subfigure}[t]{1.9in}
    \centering
    \includegraphics[height=1.45in]{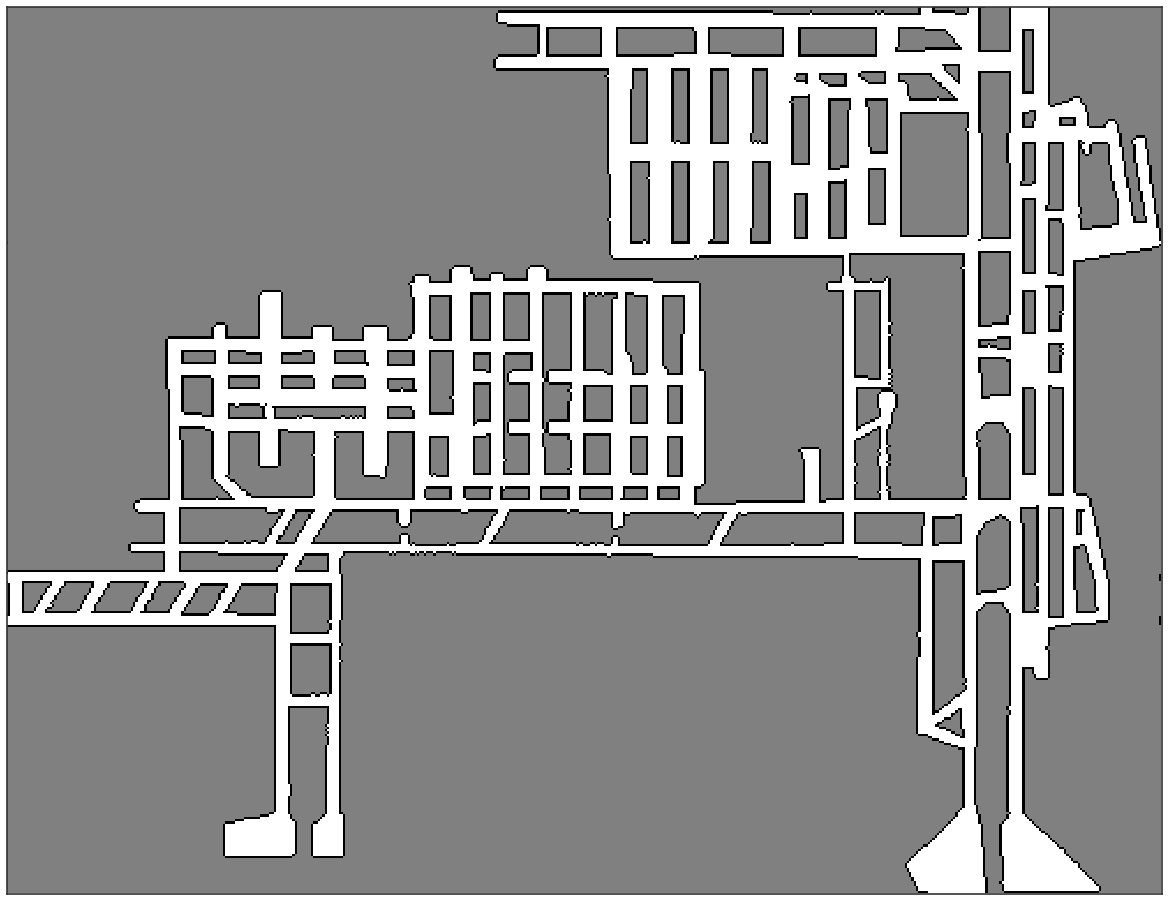}
    \caption{Tunnels ($500 \times 350$ m)}
    \label{fig:gt_tunnel}
\end{subfigure}
\hfill
\begin{subfigure}[t]{1.9in}
    \centering
    \includegraphics[height=1.45in]{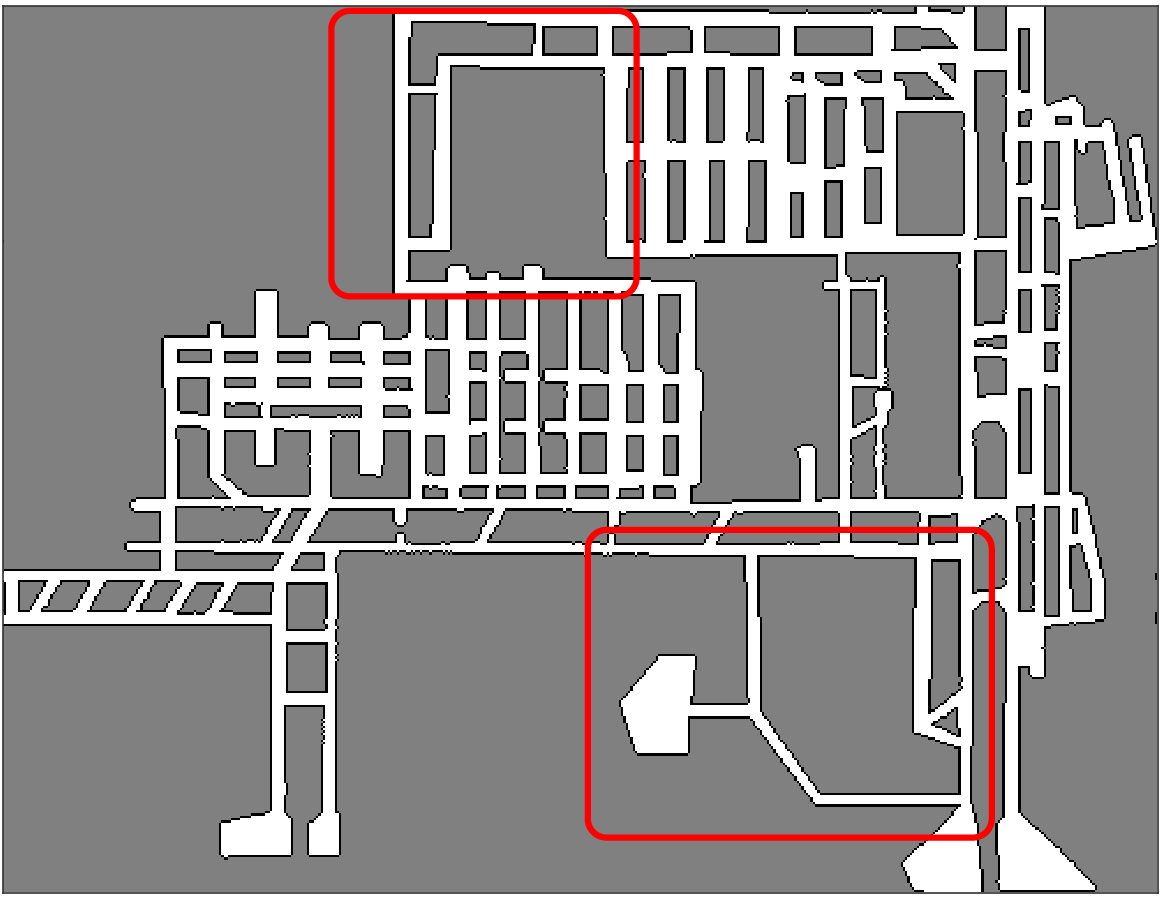}
    \caption{Modified Tunnels ($500 \times 350$ m)}
    \label{fig:gt_mod_tunnel}
\end{subfigure}
\hfill
\begin{subfigure}[t]{1.6in}
    \centering
    \includegraphics[height=1.45in]{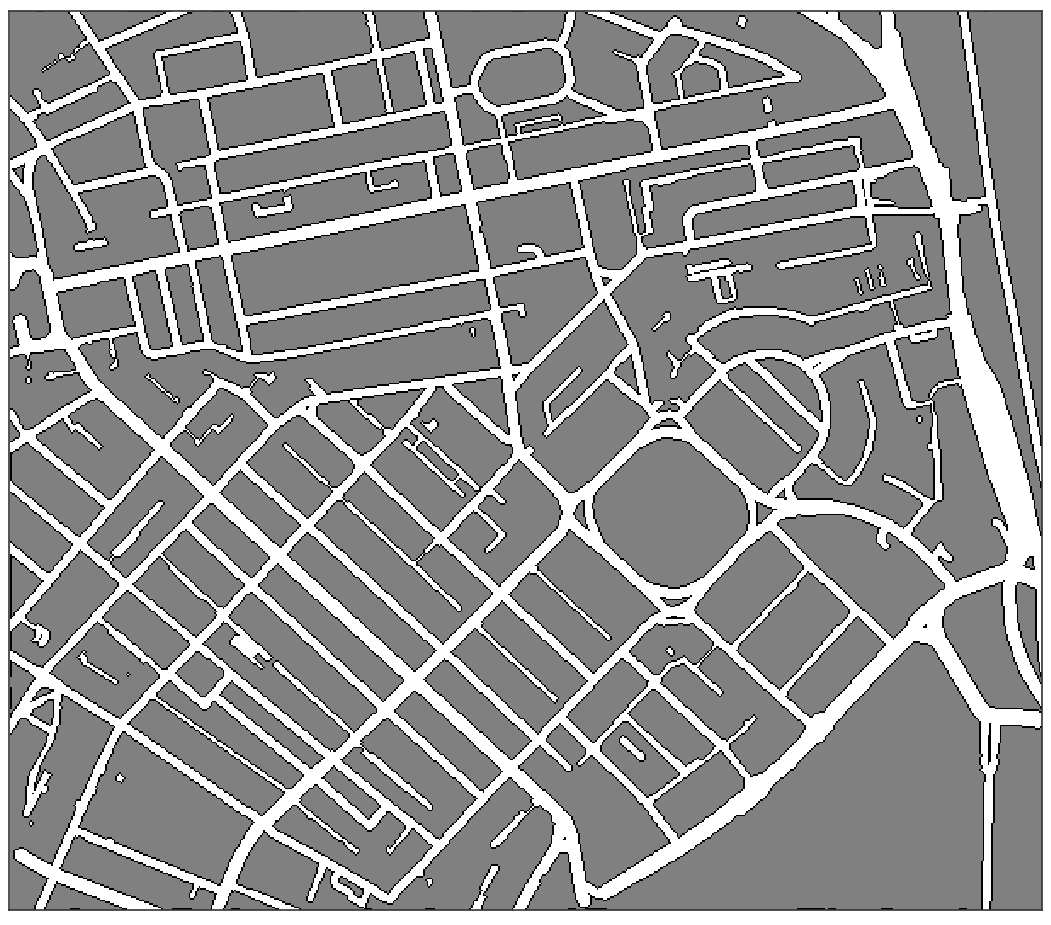}
    \caption{Urban ($600 \times 500$ m)}
    \label{fig:gt_london}
\end{subfigure}

\caption{Simulated maps used for testing communication-constrained exploration. The maps range in size and geometric structure. The red boxes indicate additions made to existing maps.}
\vspace{-0.6cm}
\label{fig:combined_maps}

\end{figure*}

\subsection{Updating Communication Points}
Algorithm \ref{alg:update_comm_points} describes the procedure for updating the communication points $\mathcal{C}_{ij}$.
The procedure first clusters the current frontiers of the fused map into exploration regions $\mathcal{Q}$, providing a coarse partition of the unexplored space. These regions are then distributed across the entire team through a process that assigns \textit{unclaimed} regions to the robot $k$ with the nearest \textit{claimed} region $q_k$, using the last known information for any robots not currently in communication. With $|\mathcal{R}|$ clusters, aggregate centroids $\bar{c}_k \ \forall k \in\mathcal{R}$ are then computed to summarize the area each robot is expected to cover. 

To facilitate efficient planning, we abstract the known free space into two dynamically updated graphs: a movement graph $G_{\text{move}}=(V,E_{\text{move}})$,  encoding traversability, and a communication graph $G_{\text{comm}}=(V,E_{\text{comm}})$, encoding which pairs of locations can establish communication. The graphs share the same vertex set $V$ but differ in their edges. For a robot pair $(i, j)$, the cluster centroids $\hat{c}_i, \hat{c}_j$ are first snapped to their nearest vertices in $V$. The communication points solver then searches for a pair of vertices $v_i, v_j\in V$ that minimizes the maximum travel distance from each centroid subject to the pair being connected in $G_{\text{comm}}$:
\begin{equation}
\begin{aligned}
\min_{v_i, v_j \in V} \quad & \max_{k \in \{i,j\}} d_{\text{move}}(\hat{c}_k, v_k) \\
\text{subject to} \quad & (v_i, v_j) \in E_{\text{comm}},
\end{aligned}
\label{eq:simple_wavefront}
\end{equation}
where $d_{\text{move}}$ denotes shortest-path distance in $G_{\text{move}}$. This is solved via an optimal wavefront search that expands outward from each centroid until a pair of vertices connected by an edge in $G_{\text{comm}}$ is found. The set $\mathcal{C}_{ij}$ is then defined by sampling points along $\overline{v_i v_j}$ such that $\texttt{inComms}()$ returns true. This set forms the spatial axis, or corridor, along which robot $i$ can plan to re-establish contact with robot $j$ at the next communication opportunity.

\subsection{Communication-Aware Exploration}
\label{subsec:comm_aware}

The communication-aware planner is formulated as a budgeted Vehicle Orienteering Problem (VOP) in which a robot must travel from a source (its current position) to a sink (any point in $\mathcal{C}_{ij}$) within a time budget $B = t^{c}_{ij} - t$. The candidate node set is the frontier set $\mathcal{F}_i$, where each frontier $f$ is assigned a prize equal to $U_{\text{front}_i}$ from Eq. ~\eqref{eq:frontier_score} with the travel distance term removed, since travel cost is already represented explicitly through edge weights in the VOP. 
The planner constructs a distance matrix over the node set $\{\text{pos}_i\} \cup \mathcal{F}_i \cup \mathcal{C}_{ij}$, using a batch of Dijkstra searches on $G_{\text{move}}$. 
The resulting prizes, distances, and budget are then passed to OR-Tools to solve for a near-optimal route. 

If no feasible route exists, typically because no communication point is reachable within $B$, the planner checks whether $\mathcal{C}_{ij}$ is reachable with an added slack time $\Delta t_\text{slack}$ such that the robot arrives $near$ $t^{c}_{ij}$. Since plans are executed in a receding-horizon fashion, this handles the case where the robot has already committed to its last viable frontier and simply needs to proceed to $\mathcal{C}_{ij}$. If $\mathcal{C}_{ij}$ is not reachable either, the planner falls back to the default exploration strategy.

Since this formulation does not include a lower-bound budget constraint, the solver may propose a chain of frontiers with an arrival to $\mathcal{C}_{ij}$ earlier than the scheduled $t^{c}_{ij}$. This is an intentional design choice. In exploration, new information is continuously revealed, so effective behavior depends on frequent replanning based on the latest available knowledge. A route that exactly consumes the available budget on the current map is unlikely to remain the best route as exploration progresses, since newly discovered regions may reveal frontiers that are more valuable than those originally selected. As a result, the planner re-solves the orienteering problem after each visited frontier. 
 
\section{Simulation and Results}

\begin{figure*}[!t]
    \centering
    
    \begin{subfigure}[b]{0.24\textwidth}
        \centering
        \includegraphics[width=\linewidth]{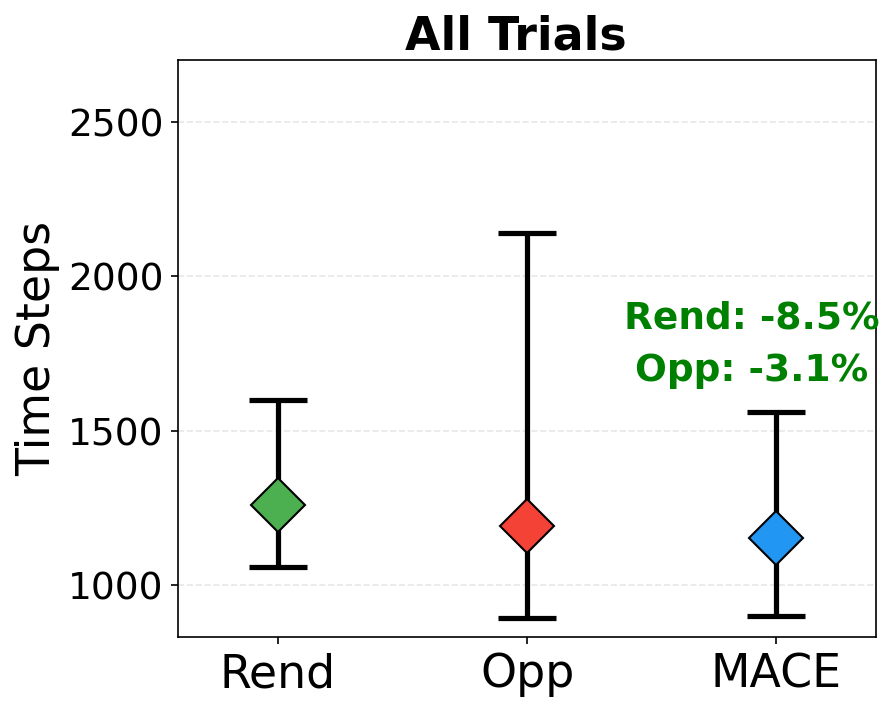}
        \label{fig:maze_all}
    \end{subfigure}
    \hfill
    \begin{subfigure}[b]{0.24\textwidth}
        \centering
        \includegraphics[width=\linewidth]{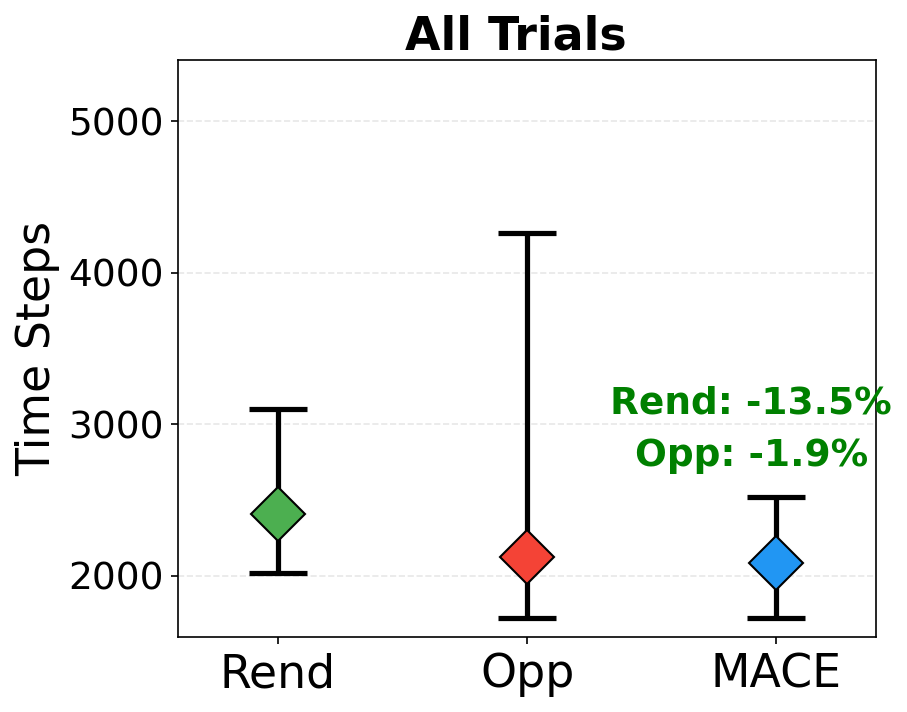}
        \label{fig:tunnel_all}
    \end{subfigure}
    \hfill
    \begin{subfigure}[b]{0.24\textwidth}
        \centering
        \includegraphics[width=\linewidth]{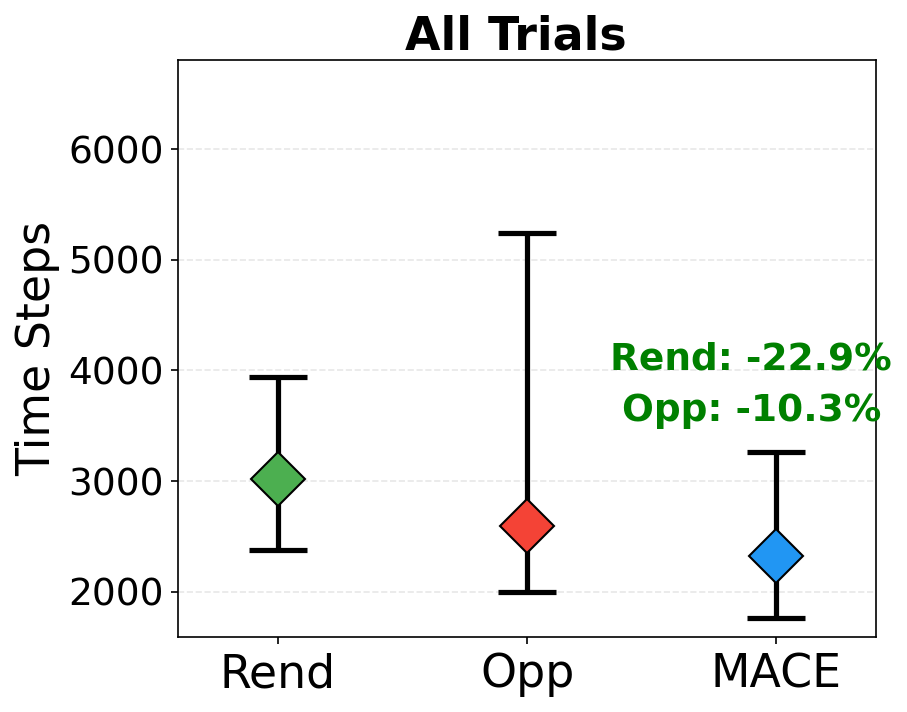}
        \label{fig:mod_tunnel_all}
    \end{subfigure}
    \hfill
    \begin{subfigure}[b]{0.24\textwidth}
        \centering
        \includegraphics[width=\linewidth]{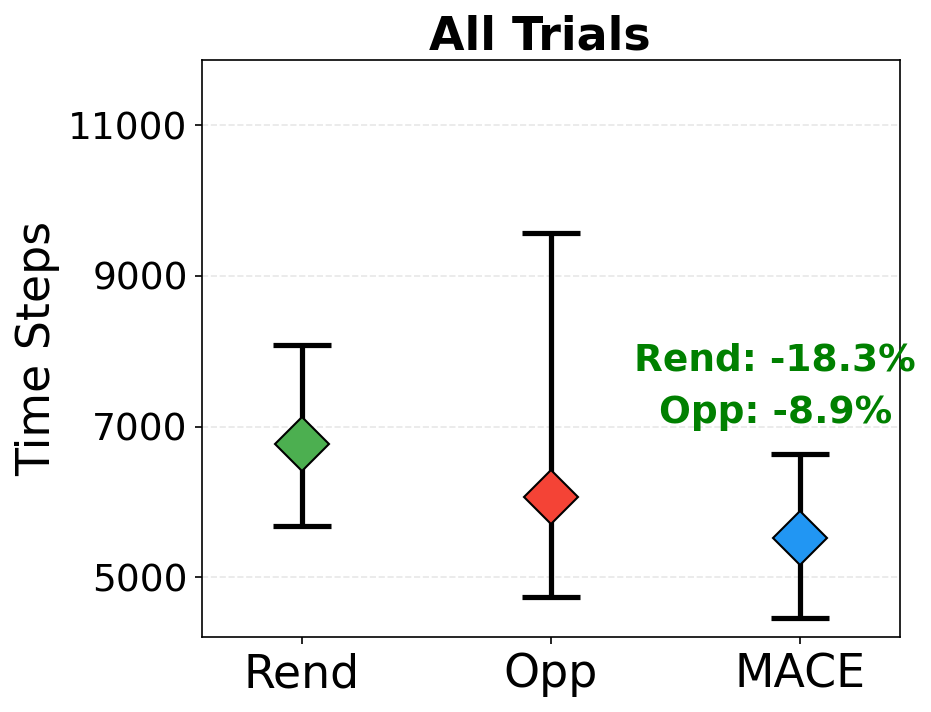}
        \label{fig:london_all}
    \end{subfigure}

    \vspace{-3pt} 

    \begin{subfigure}[b]{0.24\textwidth}
        \centering
        \includegraphics[width=\linewidth]{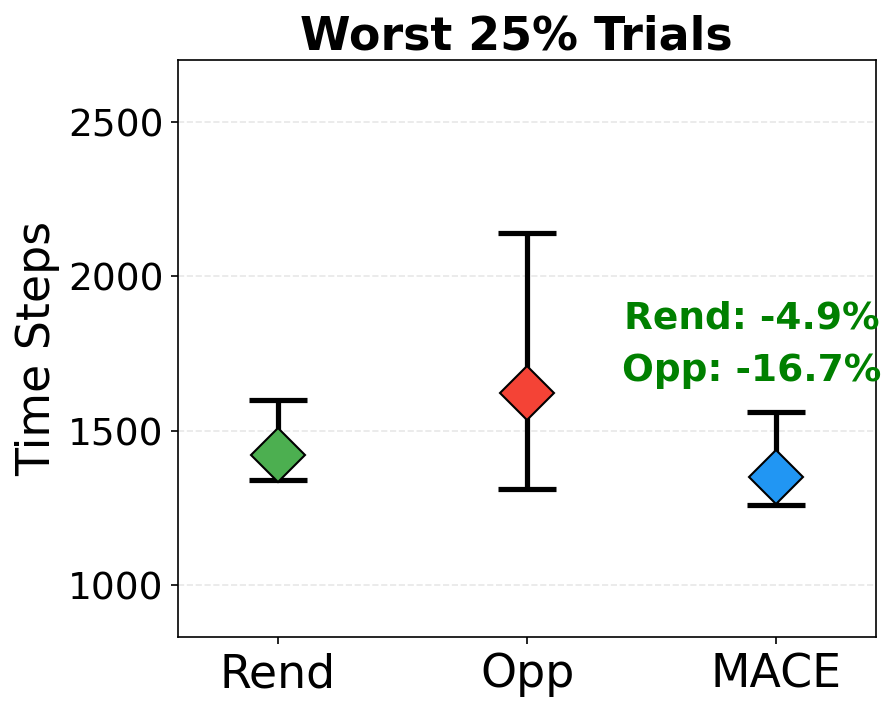}
        \caption{Maze}
        \label{fig:maze_worst}
    \end{subfigure}
    \hfill
    \begin{subfigure}[b]{0.24\textwidth}
        \centering
        \includegraphics[width=\linewidth]{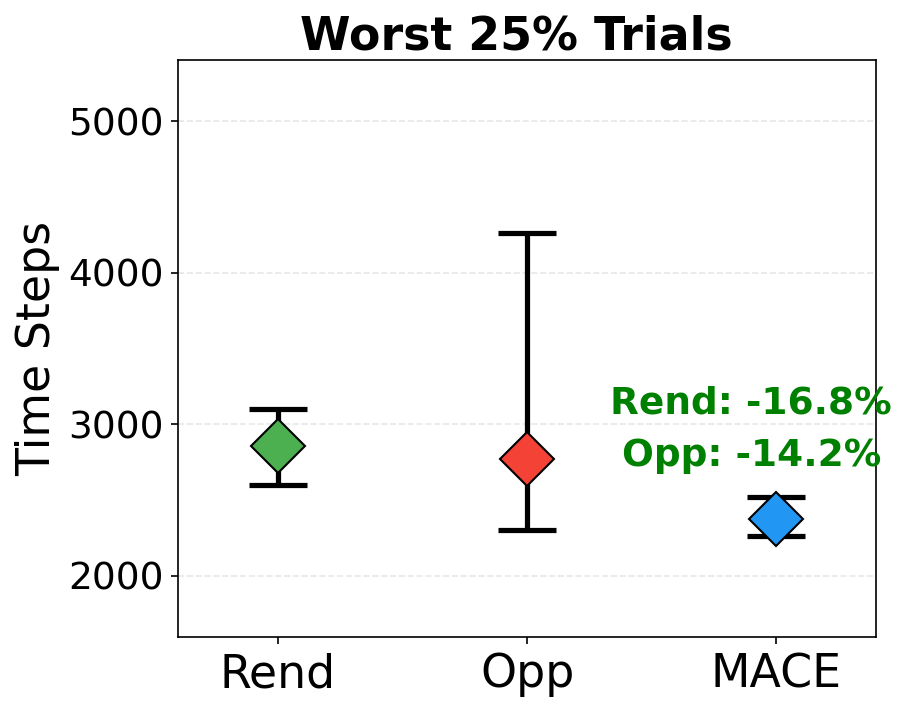}
        \caption{Tunnels}
        \label{fig:tunnel_worst}
    \end{subfigure}
    \hfill
    \begin{subfigure}[b]{0.24\textwidth}
        \centering
        \includegraphics[width=\linewidth]{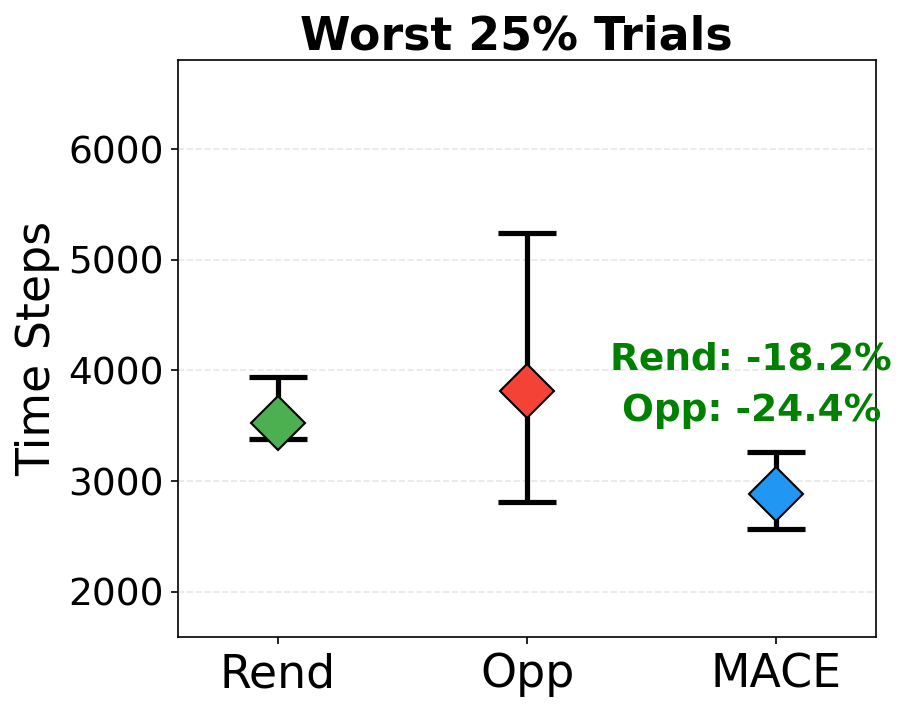}
        \caption{Modified Tunnels}
        \label{fig:mod_tunnel_worst}
    \end{subfigure}
    \hfill
    \begin{subfigure}[b]{0.24\textwidth}
        \centering
        \includegraphics[width=\linewidth]{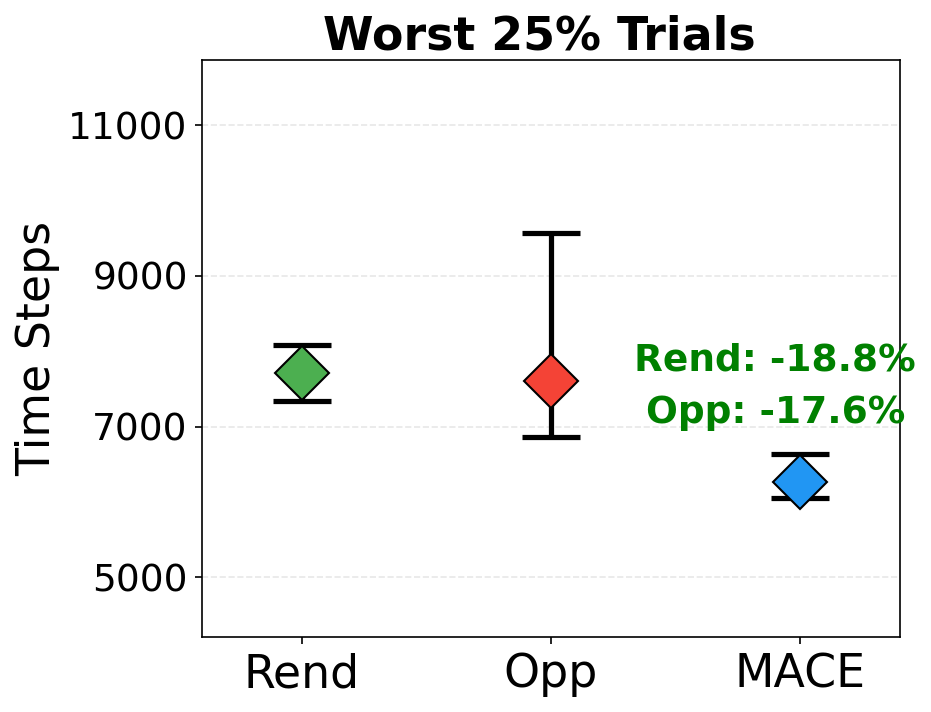}
        \caption{Urban}
        \label{fig:london_worst}
    \end{subfigure}

    \caption[Results for the total exploration time subject to long-range communication]{Results for the mean, minimum, and maximum exploration time subject to line-of-sight communication. MACE consistently outperforms the two baseline approaches, demonstrating its robustness to different types of environments and starting locations.}
    \label{fig:combined_exploration_steps_long}
    \vspace{-0.6cm}
\end{figure*}

\subsection{Simulation Setup}
To evaluate the performance of MACE, we conduct several simulated experiments across four structurally distinct environments (Fig.~\ref{fig:combined_maps}): a small-scale maze-like environment \cite{dasilva2025intermittent} ($250 \times 250$ m), a medium-scale tunnel network \cite{cao2023mtare} ($500 \times 350$ m), a modified version of the tunnel network with additional connectivity ($500 \times 350$ m), and a large-scale urban neighborhood \cite{dasilva_rendezvous} ($600 \times 500$ m). These environments vary in both size and structural connectivity (i.e., the number of interconnected passages and navigation bottlenecks). For each map, 20 starting locations are randomly sampled from free space, and each trial runs until the team achieves at least 99\% shared map coverage, defined as the fraction of free cells simultaneously known to all robots. Each trial deploys three robots with a LiDAR range of 200 m and line-of-sight (LOS) constrained communication. 

To account for the differing spatial scales of each environment, $\Delta_c$ and $\delta$ are modified to reflect the relative time required to traverse each environment. Similarly, region utility weights, frontier utility weights, and clustering parameters are adjusted proportionally to the map size. Lastly, when a robot chooses to seek communication and arrives at a communication point near the scheduled $t^{c}_{ij}$, it waits for a duration proportional to $\Delta t_\text{slack}$ before abandoning the communication point to accommodate robots that were routed to arrive at slightly different times. 

\begin{figure*}[!t]
    \centering
    
    \begin{subfigure}[b]{0.24\textwidth}
        \centering
        \includegraphics[width=\linewidth]{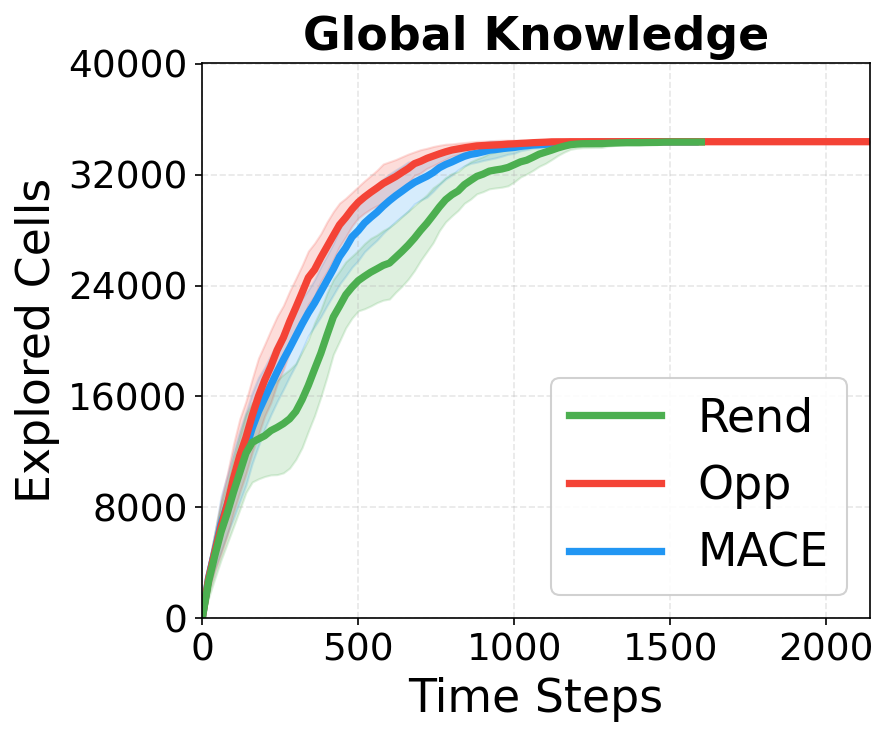}
        \label{fig:maze_union}
    \end{subfigure}%
    \hfill
    \begin{subfigure}[b]{0.24\textwidth}
        \centering
        \includegraphics[width=\linewidth]{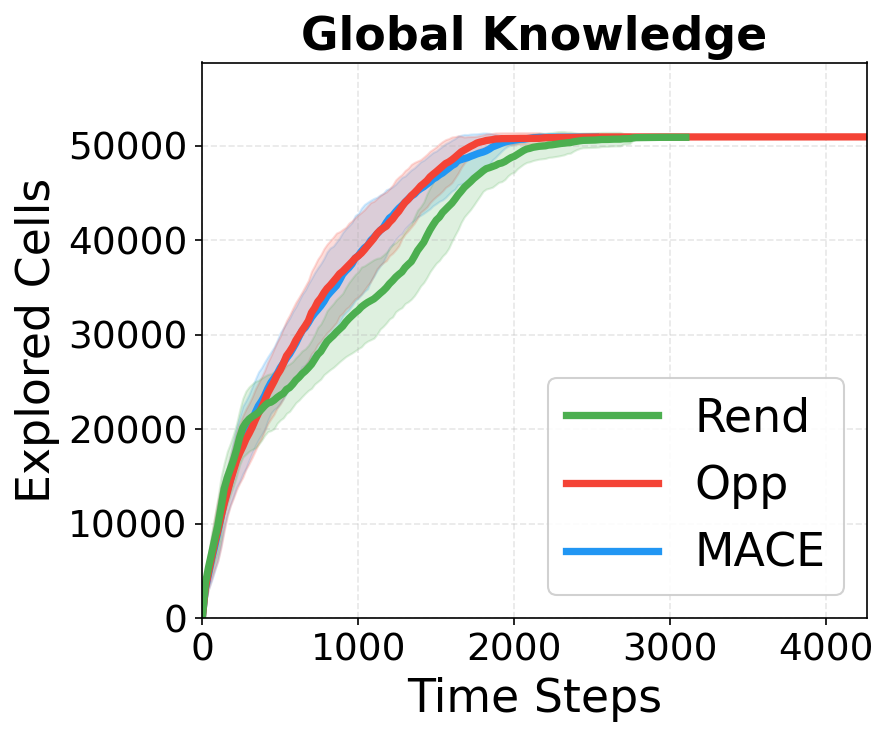}
        \label{fig:tunnel_union}
    \end{subfigure}
    \hfill
    \begin{subfigure}[b]{0.24\textwidth}
        \centering
        \includegraphics[width=\linewidth]{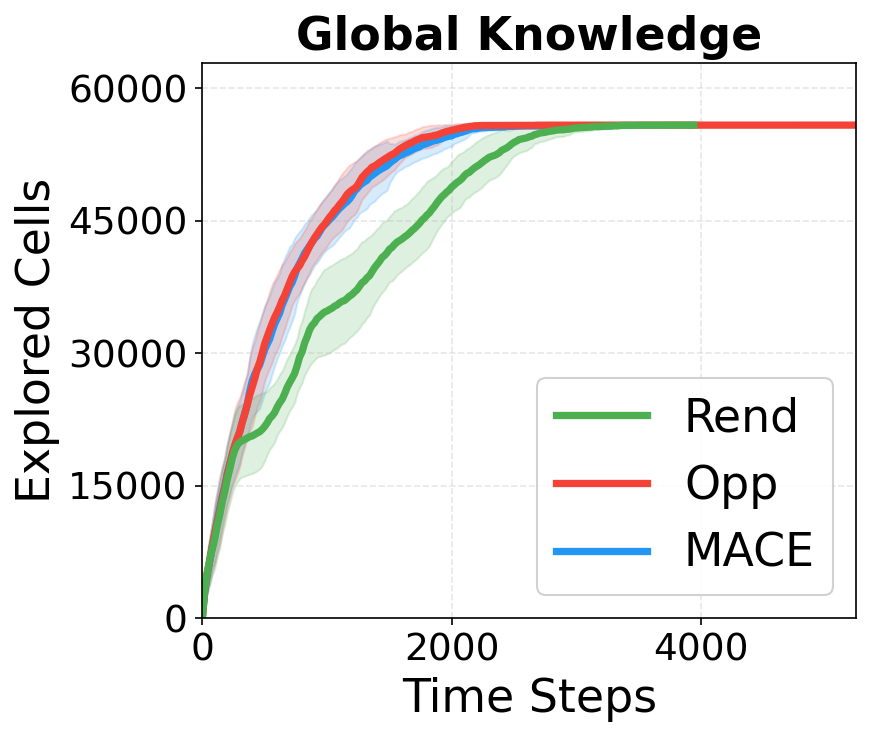}
        \label{fig:mod_tunnel_union}
    \end{subfigure}
    \hfill
    \begin{subfigure}[b]{0.24\textwidth}
        \centering
        \includegraphics[width=\linewidth]{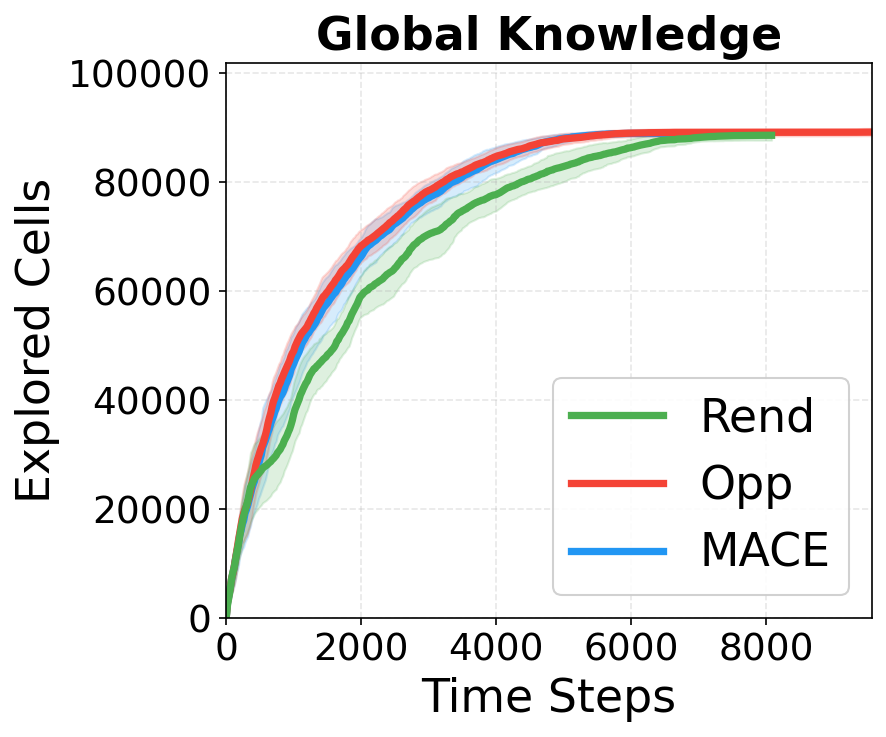}
        \label{fig:london_union}
    \end{subfigure}

    \vspace{-3pt} 

    \begin{subfigure}[b]{0.24\textwidth}
        \centering
        \includegraphics[width=\linewidth]{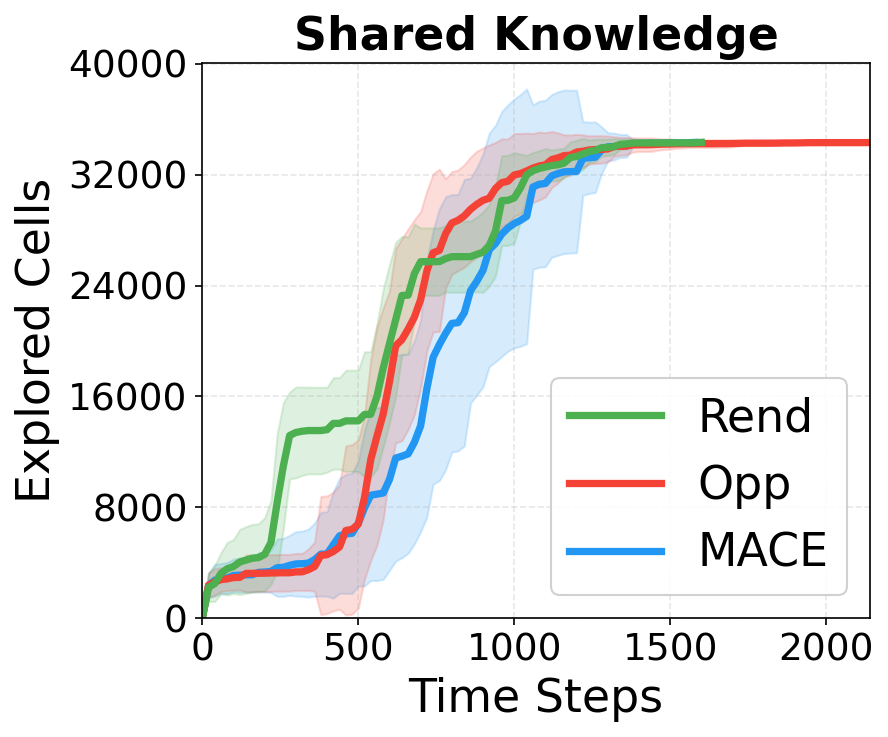}
        \caption{Maze}
        \label{fig:maze_shared}
    \end{subfigure}%
    \hfill
    \begin{subfigure}[b]{0.24\textwidth}
        \centering
        \includegraphics[width=\linewidth]{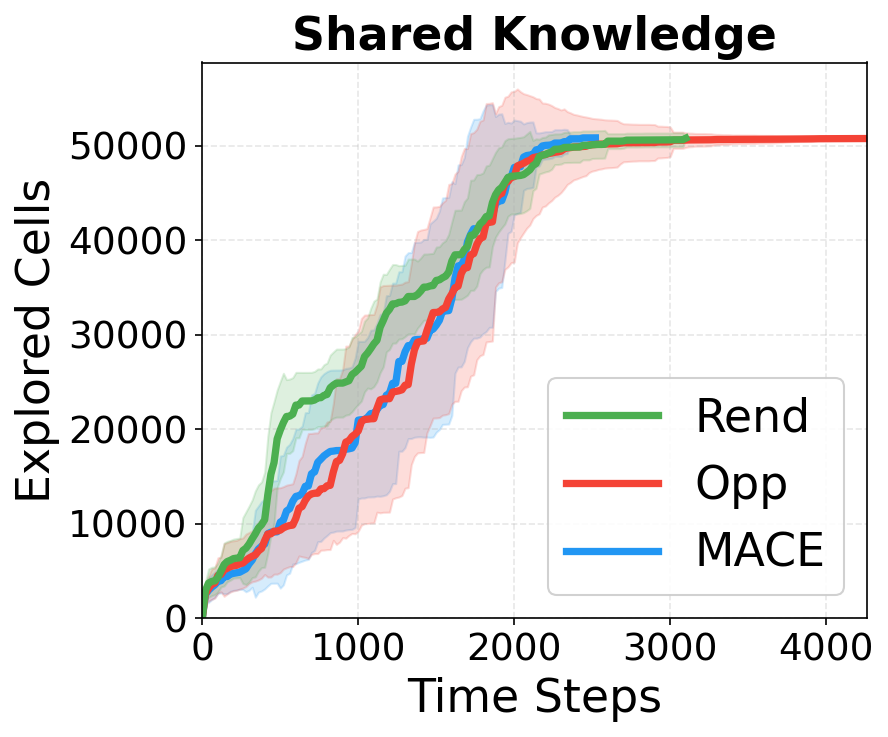}
        \caption{Tunnels}
        \label{fig:tunnel_shared}
    \end{subfigure}
    \hfill
    \begin{subfigure}[b]{0.24\textwidth}
        \centering
        \includegraphics[width=\linewidth]{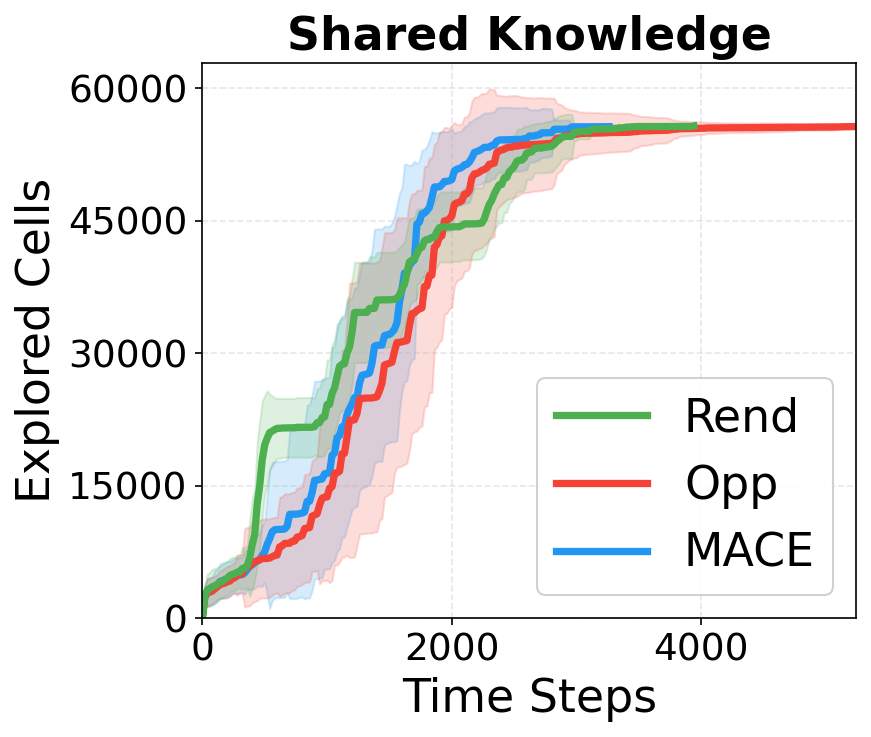} 
        \caption{Modified Tunnels}
        \label{fig:mod_tunnel_shared}
    \end{subfigure}
    \hfill
    \begin{subfigure}[b]{0.24\textwidth}
        \centering
        \includegraphics[width=\linewidth]{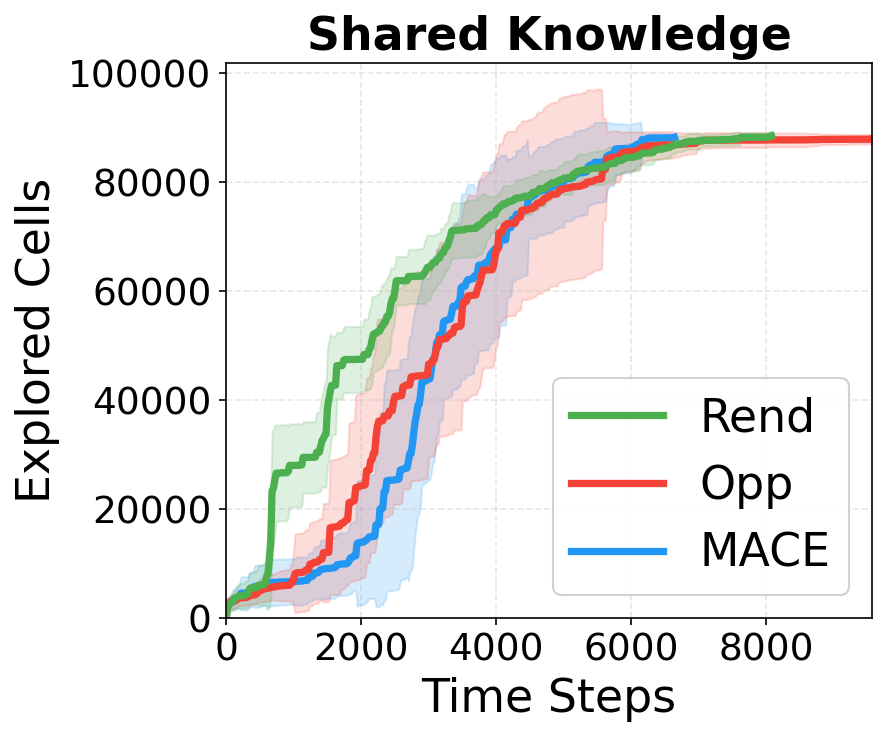} 
        \caption{Urban}
        \label{fig:london_shared}
    \end{subfigure}

    \caption[Results for expected exploration progress over time subject to short-range communication]{Expected global and shared exploration progress over time. The shaded regions represent one standard deviation from the mean.}
    \label{fig:combined_exploration_prog_long}
    \vspace{-0.6cm}
\end{figure*}

\subsection{Baselines}
We benchmark MACE against two baseline approaches:
\begin{itemize}[leftmargin=*, itemsep=2pt, topsep=2pt]
\item \textit{Rendezvous} (inspired by \cite{bramblett_rendezvous}): At each scheduled interval, the team convenes at a fixed rendezvous point defined as the center of mass of all frontier centroids observed at the previous rendezvous event. To account for travel time to and from the rendezvous location, which can be substantial in larger environments, we add a buffer to each exploration phase such that robots have approximately $\Delta_c$ time steps of true exploration between consecutive rendezvous events. During the exploration phase, this baseline uses the same default exploration approach as MACE to ensure a fair comparison of the communication strategy rather than the frontier selection policy.
\item \textit{Opportunistic}: This approach does not use any rendezvous scheduling or communication-aware navigation. Robots explore independently using the same default exploration approach as MACE and operate under the assumption that communication will occur by chance as robots come within range and LOS of one another during exploration.
\end{itemize}

\subsection{Exploration Results}

Fig. \ref{fig:combined_exploration_steps_long} depicts the mean, minimum, and maximum exploration finish times across the four maps, with each map shown for all trials (top) and the worst 25\% of trials (bottom). MACE achieves better performance than both baselines in all environments. Notably, when evaluating the worst 25\% of trials, the performance gap between MACE and the opportunistic approach increases by up to 13.6\%, demonstrating that MACE is robust to various starting locations while the opportunistic approach is more prone to redundant exploration. On the other hand, the gap between MACE and the rendezvous-based approach does not change significantly when evaluating the worst 25\% of trials. These results are in part due to MACE's fallback rendezvous mechanism after a large number of missed communication opportunities, which prevents it from degrading to the worst-case redundant exploration that the opportunistic approach suffers from.

Fig. \ref{fig:combined_exploration_prog_long} illustrates two measures of team knowledge over time: global knowledge, $E_1 \cup E_2 \cup E_3$, denoting the union of all robots' local exploration $E$ regardless of whether it has been shared, and shared knowledge, $E_1 \cap E_2 \cap E_3$, denoting the intersection of all robots' local exploration. Across all four environments, the rendezvous-based approach tends to lead in shared knowledge for much of the trial, but consistently lags in global knowledge, illustrating the coverage cost of regular backtracking to rendezvous. The curves for MACE and the opportunistic method follow a similar trajectory overall, though MACE avoids the long late-stage plateaus caused by redundant exploration.

\subsection{Map Connectivity}
\begin{figure*}[!t]
\centering

\includegraphics[width=0.85\textwidth]{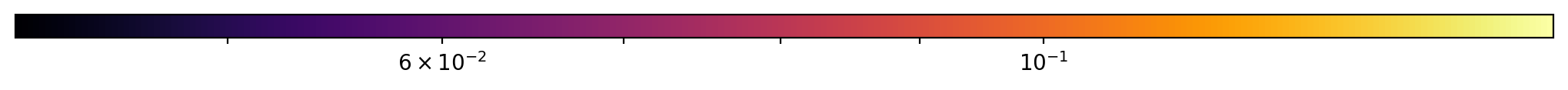}\\[5pt]
\vspace{-0.1cm}
\begin{subfigure}[t]{1.4in}
    \centering
    \includegraphics[height=1.45in]{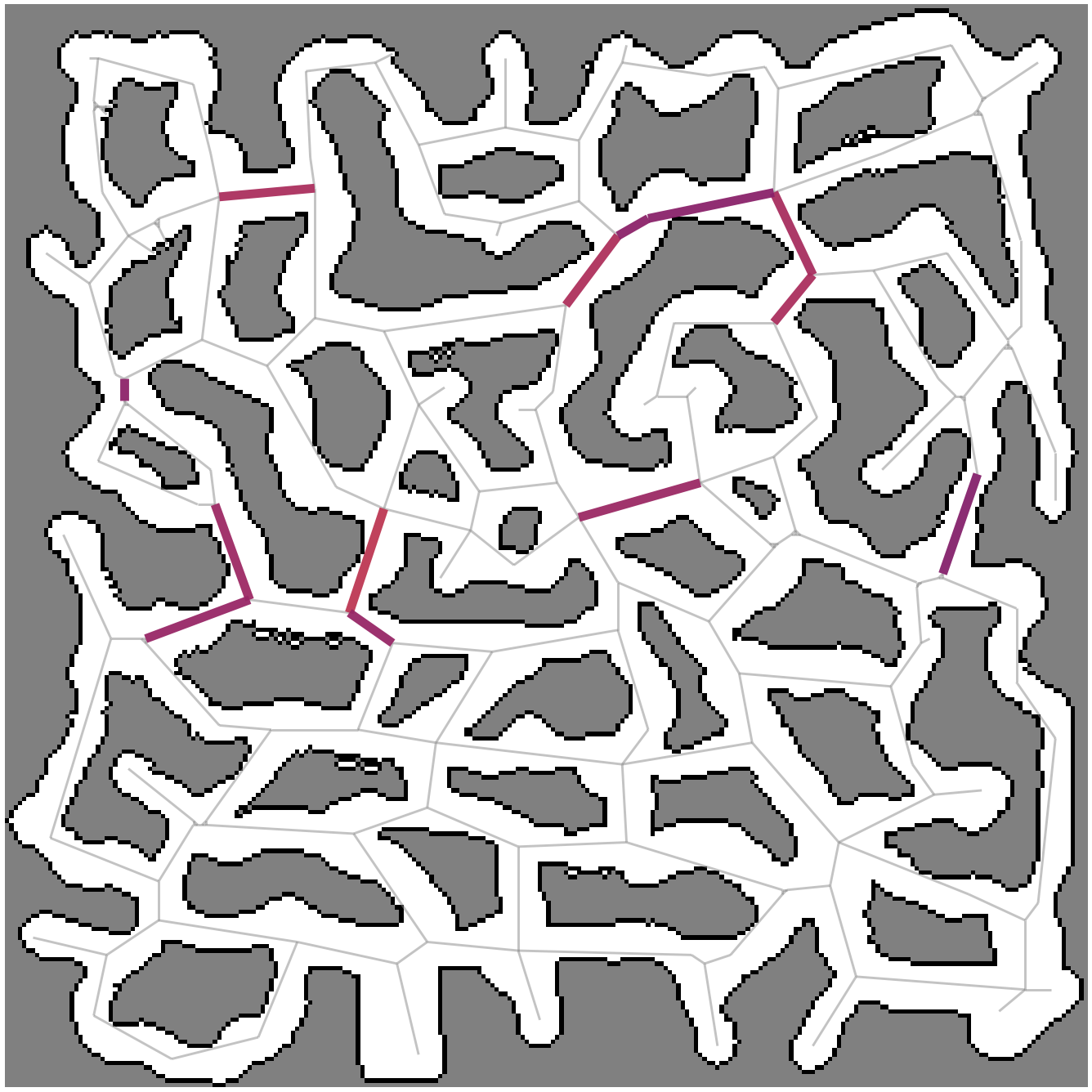}
    \caption{Maze}
    \label{fig:cfb-maze}
\end{subfigure}
\hfill
\begin{subfigure}[t]{1.9in}
    \centering
    \includegraphics[height=1.45in]{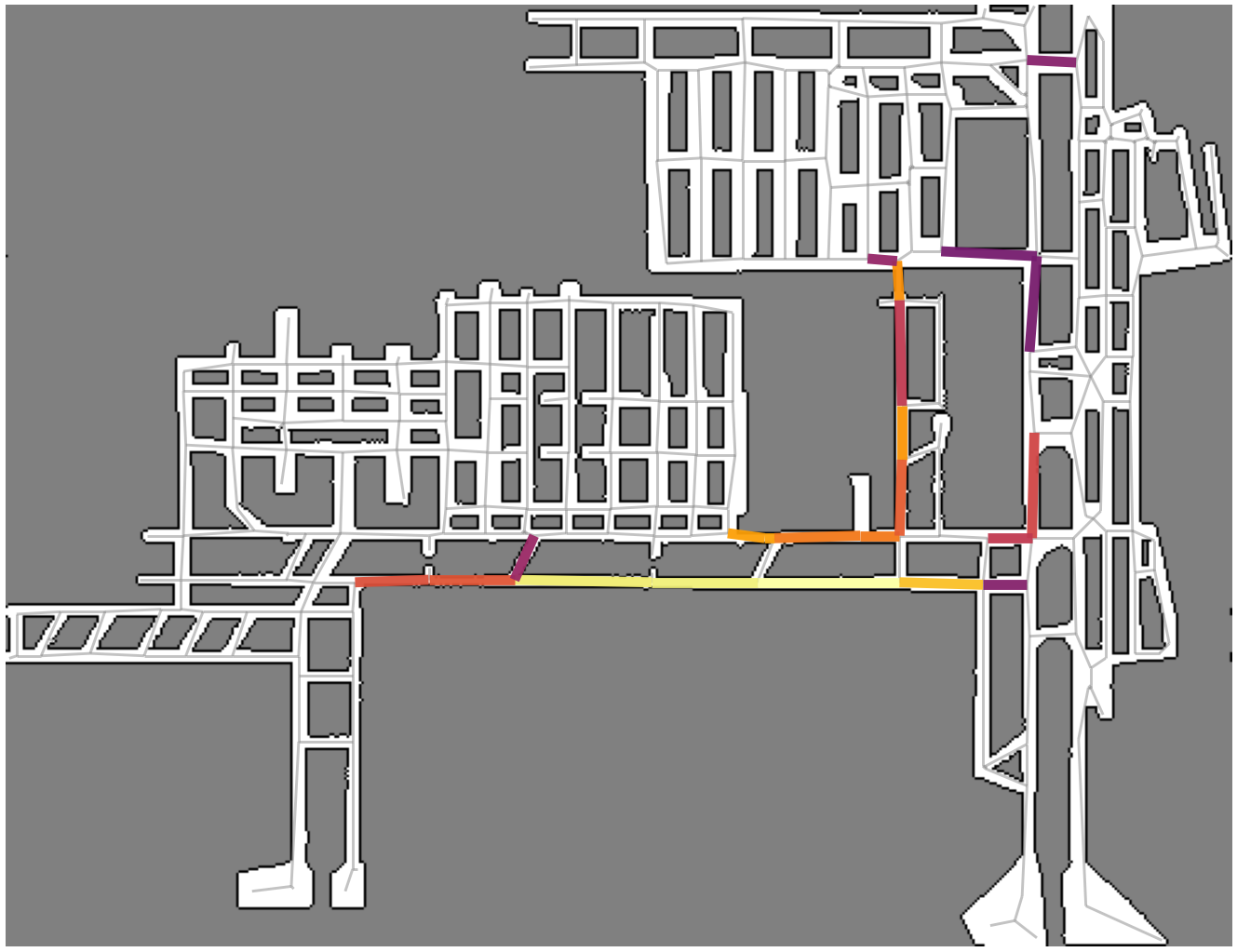}
    \caption{Tunnels}
    \label{fig:cfb-tunnel}
\end{subfigure}
\hfill
\begin{subfigure}[t]{1.9in}
    \centering
    \includegraphics[height=1.45in]{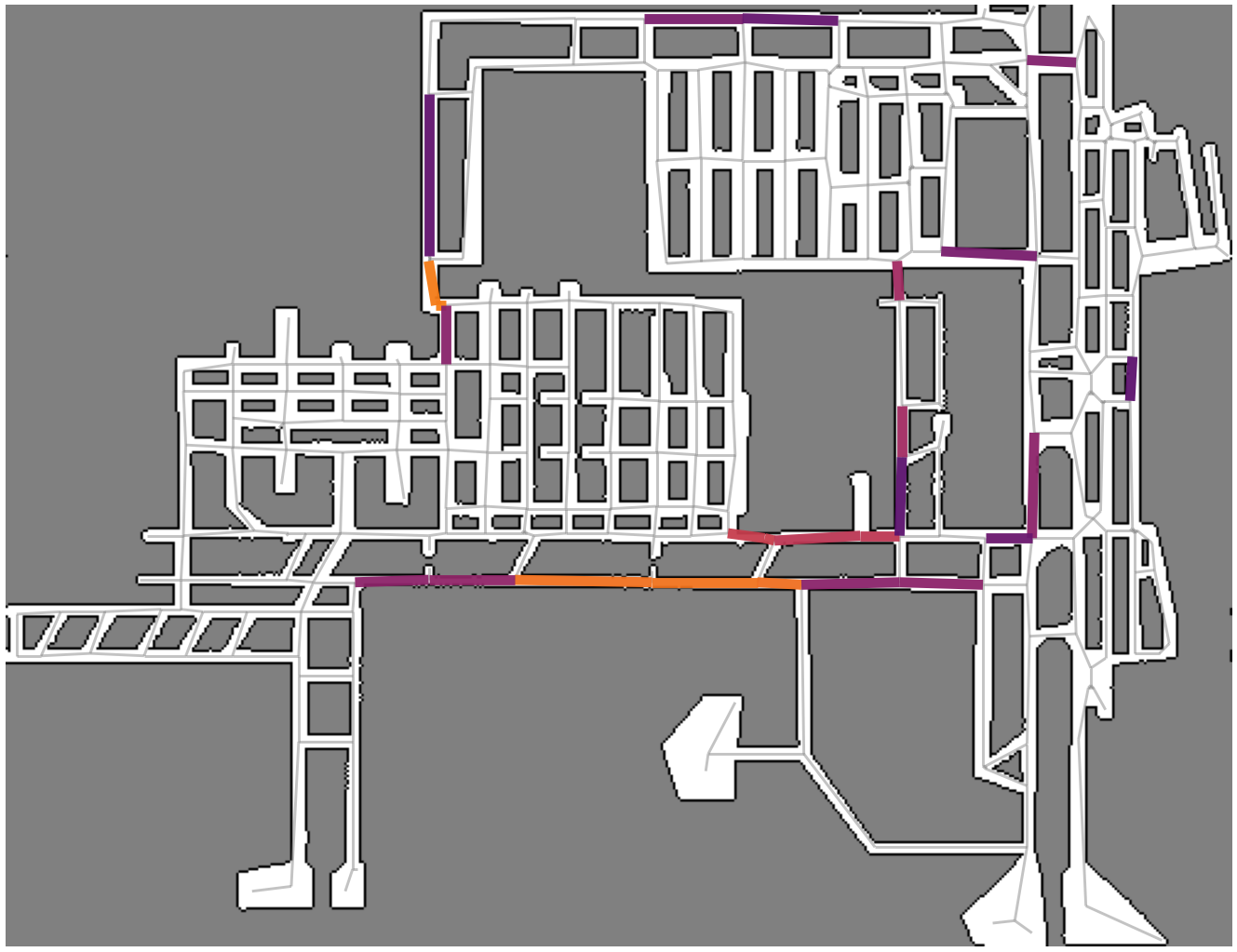}
    \caption{Modified Tunnels}
    \label{fig:cfb-tunnel-mod}
\end{subfigure}
\hfill
\begin{subfigure}[t]{1.6in}
    \centering
    \includegraphics[height=1.45in]{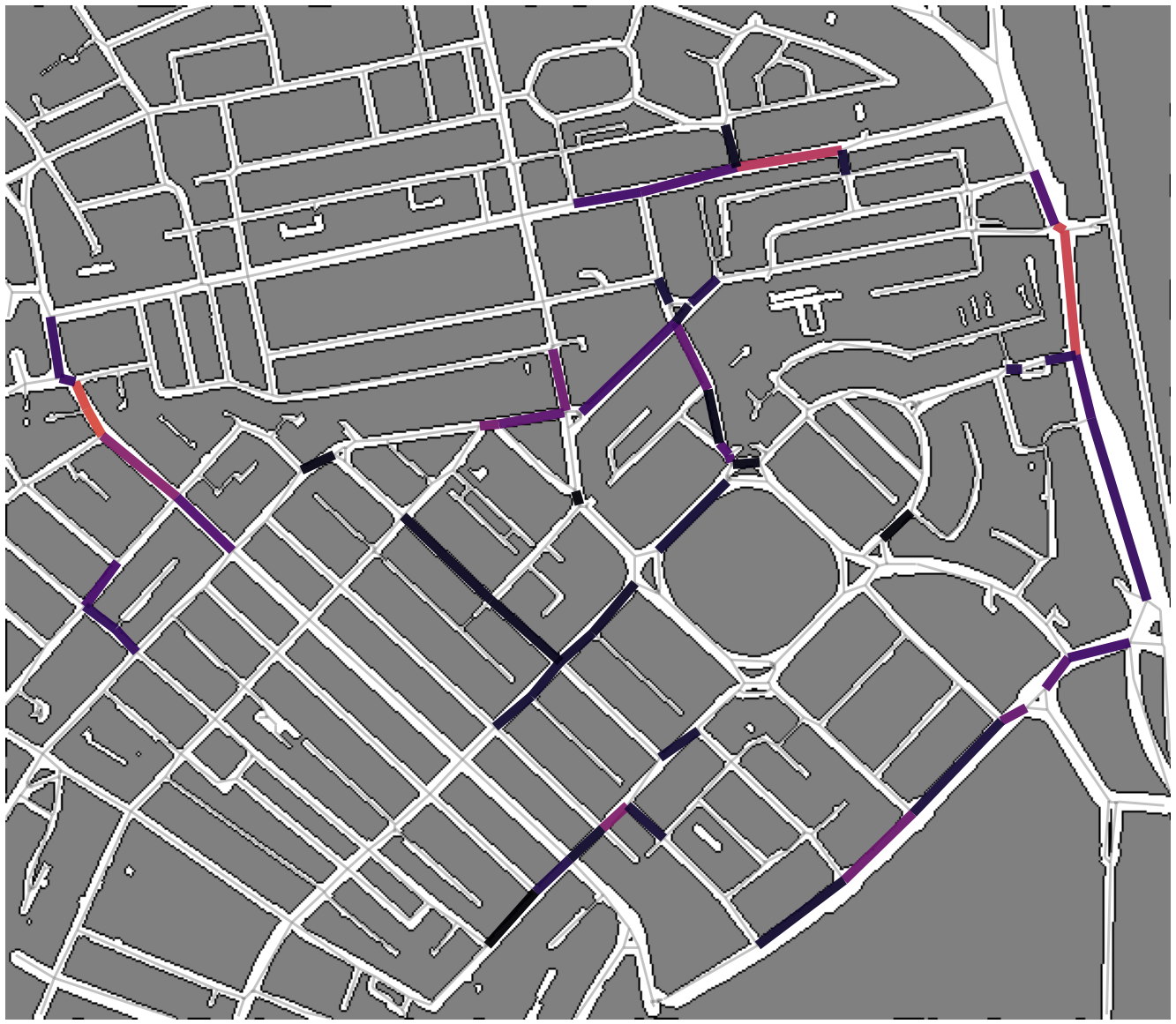}
    \caption{Urban}
    \label{fig:cfb-london}
\end{subfigure}

\caption[Top 5\% of edges ranked by current-flow betweenness for each map]{Top 5\% of edges ranked by current-flow betweenness for each map.}
\label{fig:cfb_all_maps}
\vspace{-0.3cm}
\end{figure*}

To better understand how environment structure influences the performance of different exploration strategies, we analyze each map using graph-based connectivity metrics. One such metric is current-flow betweenness (CFB) \cite{brandes2005centrality}. By treating each map as a circuit, CFB measures how much current passes through each edge, highlighting corridors and junctions that play a central role in connecting the environment. Fig. \ref{fig:cfb_all_maps} highlights the edges in the top 5\% of CFB values within each environment. From the original tunnel network to the modified tunnel network, the magnitude of CFB drops significantly because the additional connections provide alternative paths that distribute flow away from the main junctions. Likewise, the largest CFB values in both the maze and urban environments are considerably lower than those in the tunnel networks, indicating that robots generally have more directions of exploration at each junction.


CFB alone does not fully capture how environment structure influences the performance of each exploration strategy. Two additional factors, map size and communication visibility, play important roles. To account for these, we introduce an encounter probability metric derived from the communication graph. Formally,
\begin{equation}
P_{\text{encounter}}
=
\frac{|E_{\text{comm}}|}
{\binom{|V|}{2}}.
\end{equation}
$P_{\text{encounter}}$ measures the probability that any two positions in the map share a communication edge, which serves as a global metric in contrast to the local structure that CFB captures. Importantly, this metric is heavily influenced by map size. The denominator grows as $\mathcal{O}(|V|^2)$, so larger maps tend to have smaller $P_{\text{encounter}}$ even when many line-of-sight opportunities exist. This scaling is intentional as larger environments reduce the likelihood that two robots will occupy mutually visible positions at the same time, thereby increasing the risk of redundant exploration. In contrast, smaller environments inherently keep robots in closer proximity, making opportunistic encounters more common. 

\begin{figure}[t]
    \centering
    \includegraphics[width=0.35\textwidth]{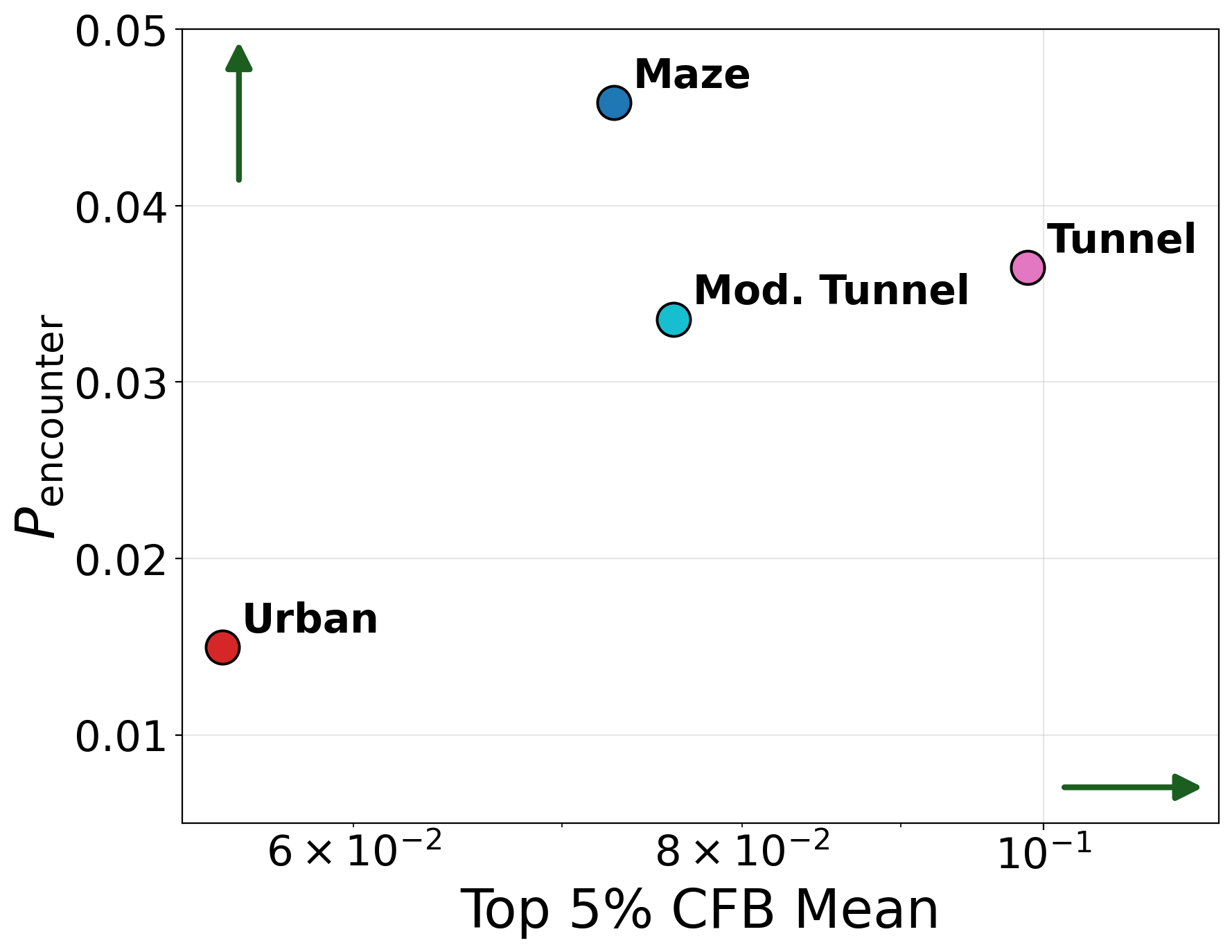}
    \vspace{-0.1cm}
    \caption[Scatter of all maps based on suitability for opportunistic exploration]{Scatter plot of the maps based on suitability for opportunistic exploration. Green arrows indicate the direction that is better suited for opportunistic exploration.}
    \label{fig:cfb_vs_p_enc}
\end{figure}

Fig. \ref{fig:cfb_vs_p_enc} plots the average CFB value of the top 5\% against
$P_{\text{encounter}}$ for each environment. Although the maze environment exhibits relatively low CFB values, its compact size increases the likelihood of opportunistic encounters between robots. Similarly, the original tunnel environment benefits from its moderate scale and prominent navigation bottlenecks, making it better suited for opportunistic communication. In contrast, the modified tunnel environment, despite having a similar size, is less favorable for opportunistic communication because the added inter-region connections reduce the dependence on a small set of corridors. Finally, the urban environment is the least conducive to opportunistic communication. Its large scale substantially decreases the probability that robots are at nearby locations, while the absence of any bottleneck corridors further reduces the likelihood of encounters.

 
These observations are generally consistent with the results previously discussed in Fig. \ref{fig:combined_exploration_steps_long}. When comparing MACE to the opportunistic approach, we see that the performance is comparable in the maze and original tunnel environments (only 3.1\% and 1.9\% mean improvement, respectively). However, in the modified tunnel and urban environments, the mean performance improves by 10.3\% and 8.9\%, respectively, reflecting the implications of larger and more connected environments. While these trends do not mimic the trends in Fig. \ref{fig:combined_exploration_steps_long}, they suggest that the proposed metrics capture several of the key structural differences between the environments. 

\section{Conclusion}
In this work, we introduced MACE, a decentralized framework for communication-aware multi-robot exploration that actively incorporates communication-seeking behavior into team coordination. Our results demonstrate that MACE consistently achieves the lowest average exploration time across a set of diverse maps, ranging in size and geometry. Furthermore, we highlight the relationship between exploration strategy and environment structure through two graph-based metrics, opening up a promising avenue for improved exploration strategies that are conditioned on the map geometry in real time. Future work includes expanding the simulation environment beyond grid-world experiments to a higher fidelity simulator such as Gazebo. This would enable testing against other state-of-the-art methods, like the pursuit-based M-TARE planner \cite{cao2023mtare}.



\bibliographystyle{IEEEtran}
\bibliography{references}
\addtolength{\textheight}{-12cm}   





\end{document}